\documentclass{article} %
\usepackage[table,xcdraw]{xcolor}
\usepackage{iclr2021_conference,times}

\usepackage{amsmath,amsfonts,bm}

\def\eqref#1{equation~\ref{#1}}

\def\1{\bm{1}}

\DeclareMathAlphabet{\mathsfit}{\encodingdefault}{\sfdefault}{m}{sl}
\SetMathAlphabet{\mathsfit}{bold}{\encodingdefault}{\sfdefault}{bx}{n}

\usepackage[hidelinks]{hyperref}
\usepackage{url}
\usepackage{inconsolata}
\usepackage[utf8]{inputenc} %
\usepackage[T1]{fontenc}    %
\usepackage{url}            %
\usepackage{booktabs}       %
\usepackage{amsfonts}       %
\usepackage{nicefrac}       %
\usepackage{microtype}      %
\usepackage{dsfont}
\usepackage{textcomp}
\usepackage{tikz}
\usepackage{multirow}
\usetikzlibrary{shapes}
\usetikzlibrary{arrows}
\usetikzlibrary{positioning}
\usetikzlibrary{bayesnet}
\usepackage[separate-uncertainty=true]{siunitx}
\usepackage{amsmath}
\usepackage{enumitem}
\usepackage{amssymb}
\usepackage{scalefnt}
\usepackage{graphicx}
\usepackage{pdfpages}
\usepackage{wrapfig}
\usepackage{stmaryrd}
\usepackage{pmboxdraw}
\usepackage{bbold}
\usepackage{xspace}
\usepackage{comment}
\usepackage[capitalize]{cleveref}
\graphicspath{ {./figures/} }
\usepackage{cleveref}
\usepackage{subfig}

\newcommand\ourmethod{\textsc{R\&R}\xspace}

\title{Learning to Recombine and Resample Data for Compositional Generalization}
\newcommand{\stderr}[1]{\scriptsize $\pm #1$}
\usepackage[font=small]{caption}
\usepackage{float}

\newcommand\prewrite{p_\textrm{rewrite}}
\newcommand\precomb{p_\textrm{recomb}}
\newcommand\defeq{\stackrel{\text { \tiny def }}{=}}
\newcommand\expect{\mathbb{E}}

\newcommand\boxRight{\textSFii\pmboxdrawuni{2574}}
\newcommand\boxDownRight{\textSFviii\pmboxdrawuni{2574}}
\newcommand\boxSpace{\hspace{1.5em}}
\newcommand{\CC}{\cellcolor{gray!15}}
\newcolumntype{L}{>{\hspace*{-\tabcolsep}}c}
\crefformat{equation}{Eq.~#2#1#3}

\renewcommand{\paragraph}[1]{\textbf{#1}~}
\newcommand*\rot{\rotatebox{90}}

\author{%
  Ekin Akyürek  \\
  MIT CSAIL\\
  \texttt{akyurek@mit.edu} \\
  \And
  Afra Feyza Akyürek  \\
  Boston University\\
  \texttt{akyurek@bu.edu} \\
  \And
  Jacob Andreas \\
   MIT CSAIL\\
  \texttt{jda@mit.edu} \\
  
}

\iclrfinalcopy

\begin{document}
\maketitle

\begin{abstract}
Flexible neural sequence models outperform grammar- and automaton-based counterparts on a variety of tasks. However, neural models perform poorly in settings requiring compositional generalization beyond the training data---particularly to rare or unseen 
subsequences. Past work has found symbolic scaffolding (e.g.\ grammars or automata) essential in these settings. We describe \ourmethod, a \emph{learned data augmentation} scheme that enables a large category of compositional generalizations without appeal to latent symbolic structure.
\ourmethod has two components: \emph{recombination} of original training examples via a prototype-based generative model and \emph{resampling} of generated examples to encourage extrapolation.
Training an ordinary neural sequence model on a dataset augmented with recombined and resampled examples significantly improves generalization in two language processing problems---instruction following (\textsc{scan}) and morphological analysis (\textsc{sigmorphon} 2018)---where \ourmethod enables learning of new constructions and tenses from as few as eight initial examples.

\vspace{-0.5em}
\end{abstract}

\section{Introduction}\label{sec:introduction}

\begin{wrapfigure}{r}{0.4\textwidth}
  \vspace{-1em}
  \centering
 \includegraphics[width=0.4\textwidth]{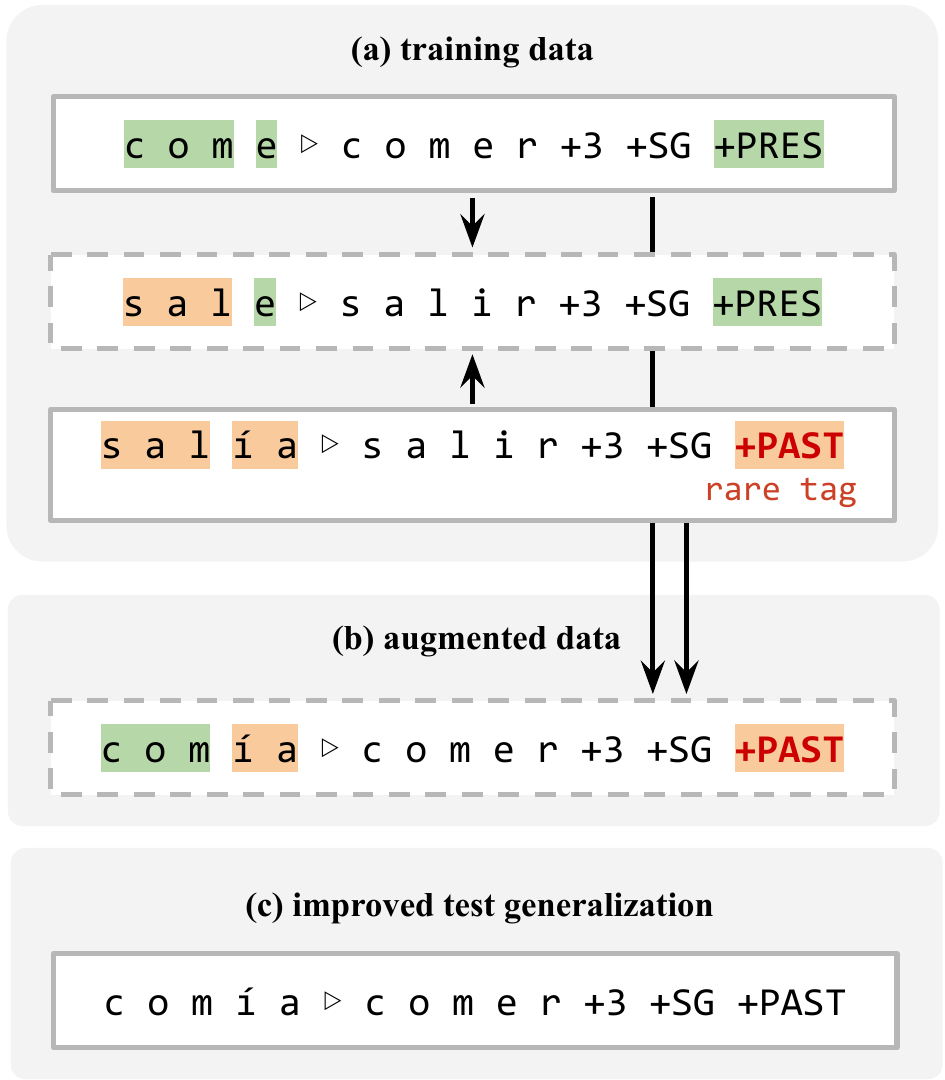} 
 \caption{We first train a generative model to reconstruct training pairs $(x \triangleright y)$ by constructing them from other training pairs (a). We then perform data augmentation by \emph{sampling} from this model, preferentially generating samples in which $y$ contains rare tokens or substructures (b). Dashed boxes show prediction targets. Conditional models trained on the augmented dataset accurately predict outputs $y$ from new inputs $x$ requiring compositional generalization (c).}
 \label{fig:teaser2}
 \vspace{-1.5em}
\end{wrapfigure}

How can we build machine learning models with the ability
to learn new concepts in context from little data?
Human language learners acquire new word meanings from a single exposure \citep{carey1978acquiring}, and immediately incorporate words and their meanings productively and compositionally into larger linguistic and conceptual systems \citep{berko1958child, piantadosi2016compositional}.
Despite the remarkable success of neural network models on many learning problems in recent years---including one-shot learning of classifiers and policies \citep{santoro2016meta, wang2016meta}---this kind of few-shot learning of composable concepts remains beyond the reach of standard neural models in both diagnostic and naturalistic settings \citep{SCAN, bahdanau2019closure}.

Consider the \emph{few-shot morphology learning} problem shown in \cref{fig:teaser2}, in which a learner must predict various linguistic features (e.g. \texttt{3}rd person, \texttt{S}in\texttt{G}ular, \texttt{PRES}ent tense) from word forms, 
with
only a small number of examples of the \texttt{PAST} tense 
in the training set.
Neural sequence-to-sequence models \citep[e.g.][]{attention-bahdanau} trained on this kind of imbalanced data fail to predict past-tense tags on held-out inputs of any kind (\cref{sec:datasetsexps}). 
Previous attempts to address this and related shortcomings in neural models have focused on explicitly encouraging rule-like behavior by e.g.\ modeling data with symbolic grammars \citep{data-recombination-copy, structuredsemparse, cai-recursion} or applying rule-based data augmentation \citep{arxiv:geca}.
These procedures involve highly task-specific models or generative assumptions, preventing them from generalizing effectively to less structured problems that combine rule-like and exceptional behavior. More fundamentally, they fail to answer the question of whether explicit rules are necessary for compositional inductive bias, and whether it is possible to obtain ``rule-like'' inductive bias \emph{without} appeal to an underlying symbolic generative process.

This paper describes a procedure for improving few-shot compositional generalization in neural sequence models without symbolic scaffolding.
Our key insight is that 
even fixed, imbalanced training datasets provide a rich source of supervision for few-shot learning of concepts and composition rules. In particular, we propose a new class of prototype-based neural sequence models \citep[c.f.][]{gu2018search} that can be directly trained to perform the kinds of generalization exhibited in \cref{fig:teaser2}
by explicitly \textbf{recombining} fragments of training examples to reconstruct other examples.
Even when these prototype-based models are not effective as general-purpose predictors,
we can \textbf{resample} their outputs to select high-quality synthetic examples of rare phenomena.
Ordinary neural sequence models may then be trained on datasets augmented with these synthetic examples, distilling the learned regularities into more flexible predictors. This procedure, which we abbreviate \textbf{\ourmethod},
promotes efficient generalization in both challenging synthetic sequence modeling tasks \citep{SCAN} and morphological analysis in multiple natural languages \citep{sigmorphon2018}.

  By directly optimizing for the kinds of generalization that symbolic representations are supposed to support, we can bypass the need for symbolic representations themselves: \ourmethod gives performance comparable to or better than state-of-the-art neuro-symbolic approaches on tests of compositional generalization.

Our results suggest that some failures of systematicity in neural models can be 
explained by simpler structural constraints on data distributions and
corrected with weaker inductive bias than previously described.\footnote{
Code for all experiments in this paper is available at \url{https://github.com/ekinakyurek/compgen}.
We implemented our experiments in Knet \citep{yuret2016knet} using Julia \citep{bezanson2017julia}.}

\section{Background and related work}\label{sec:background}

\paragraph{Compositional generalization}
Systematic compositionality---the capacity to identify rule-like regularities from limited data and generalize these rules to novel situations---is an essential feature of human reasoning \citep{fodor1988connectionism}.
While details vary, a common feature of existing attempts to formalize systematicity 
in sequence modeling problems
\citep[e.g.][]{gordon2020permutation} is the intuition that learners should make accurate predictions in situations featuring novel combinations of previously observed input or output subsequences. For example, learners should generalize from actions seen in isolation to more complex commands involving those actions \citep{lake2019human}, and from relations of the form \texttt{r(a,b)} to \texttt{r(b,a)} \citep{keysers2019measuring,bahdanau2018systematic}.
In machine learning, previous studies have found that standard neural architectures fail to generalize systematically even when they achieve high in-distribution accuracy in a variety of settings \citep{SCAN, Bastings:etal:2018, johnson2017clevr}.

\paragraph{Data augmentation and resampling}
Learning to predict sequential outputs with rare or novel subsequences is related to the widely studied problem of \emph{class imbalance} in classification problems. 
There, undersampling of the majority class or oversampling of the minority class has been found to improve the quality of predictions for rare phenomena \citep{japkowicz2000learning}.
This can be combined with targeted \emph{data augmentation} with synthetic examples of the minority class \citep{chawla2002smote}. 
Generically, given a training dataset $\mathcal{D}$, learning with class resampling and data augmentation involves defining an augmentation distribution $\tilde{p}(x, y \mid \mathcal{D})$ and sample weighting function $u(x, y)$ and maximizing a training objective of the form:
\begin{equation}
    \label{eq:objective}
    \mathcal{L}(\theta) = \underbrace{\textstyle \frac{1}{|\mathcal{D}|} \sum_{x \in \mathcal{D}} \log p_\theta(y \mid x)}_\text{Original training data} + \underbrace{\vphantom{\textstyle\frac{1}{|\mathcal{D}|}} \expect_{(x, y) \sim \tilde{p}} ~ u(x, y) \log p_\theta(y \mid x)}_\text{Augmented data} ~ .
\end{equation}

In addition to
task-specific model architectures \citep{andreas2016neural,russin2019compositional}, recent years have seen a renewed interest in data augmentation as a flexible and model-agnostic tool for encouraging controlled generalization \citep{ratner2017learning}. 
Existing proposals for sequence models are mainly rule-based---in sequence modeling problems, specifying a synchronous context-free grammar \citep{data-recombination-copy} or string rewriting system \citep{arxiv:geca} to generate new examples. Rule-based data augmentation schemes that recombine multiple training examples have been proposed for image classification \citep{dataaugmentationbypairing} and machine translation \citep{dataaugmentationlowresourcemt}.
While rule-based data augmentation is highly effective in structured problems featuring crisp correspondences between inputs and outputs, the effectiveness of such approaches involving more complicated, context-dependent relationships between inputs and outputs has not been well-studied.

\paragraph{Learned data augmentation}
What might compositional data augmentation look like without rules as a source of inductive bias? As \cref{fig:teaser2} suggests, an ideal data augmentation procedure ($\tilde{p}$ in \cref{eq:objective}) should automatically identify valid ways of transforming and combining examples, without pre-committing to a fixed set of transformations.\footnote{As a concrete example of the potential advantage of learned data augmentation,
consider applying the \textsc{geca} procedure of \citet{arxiv:geca} to the language of strings $a^n b^n$.
\textsc{geca} produces a training set that is substitutable in the sense of \citet{clark-substitutable}; as noted there, $a^n b^n$ is not substitutable. \textsc{geca} will infer that $a$ can be replaced with $aab$ based on their common context in $(a\underline{a}bb, a\underline{aab}bb)$, then generate the malformed example $a\underline{aab}abbb$ by replacing an $a$ in the wrong position. In contrast, recurrent neural networks can accurately model $a^n b^n$ \citep{weiss2018practical, gers2001lstm}. Of course, this language can also be generated using even more constrained procedures than \textsc{geca}, but in general learned sequence models can capture a broader set of both formal regularities and exceptions compared to rule-based procedures.}
A promising starting point is provided by 
\textbf{prototype-based} models, a number of which \citep{gu2018search, neural-editor, khandelwal2019generalization} have been recently proposed for sequence modeling. 
Such models generate data according to:
\begin{align}
     d \sim \prewrite( \cdot \mid d' ; \theta)  \quad \textrm{where} \quad d'\sim \textrm{Unif}(\mathcal{D}); \label{eq:generate}
\end{align}
for a dataset $\mathcal{D}$ and 
a learned sequence rewriting model $\prewrite(d \mid d'; \theta)$. 
(To avoid confusion, we will use the symbol $d$ to denote a \emph{datum}. Because a data augmentation procedure must produce complete input--output examples, each $d$ is an $(x, y)$ pair for the conditional tasks evaluated in this paper.)
While recent variants implement $\prewrite$ with neural networks, these models are closely related to classical kernel density estimators \citep{rosenblatt1956remarks}. But additionally---building on the motivation in \cref{sec:introduction}---they may be viewed as \emph{one-shot learners} trained to generate new data $d$ from a single example.

Existing work uses prototype-based models as replacements for standard sequence models. 
We will show here that they are even better suited to use as data augmentation procedures: they can produce high-precision examples in the neighborhood of existing training data, then be used to bootstrap simpler predictors that extrapolate more effectively.
But our experiments will also show that existing prototype-based models
give mixed results on challenging generalizations of the kind depicted in \cref{fig:teaser2} when used for either direct prediction or data augmentation---performing well in some settings but barely above baseline in others.

Accordingly, \ourmethod is built on two model components that transform prototype-based language models into an effective learned data augmentation scheme.
\cref{sec:prototypes} describes an implementation of $p_\textrm{rewrite}$ that encourages greater sample diversity and well-formedness via a multi-prototype copying mechanism (a \emph{two-shot} learner). \cref{sec:sampling} describes heuristics for sampling prototypes $d'$ and model outputs $d$ to focus data augmentation on the most informative examples. \cref{sec:datasetsexps} investigates the empirical performance of both components of the approach, finding that
they together provide they a simple but surprisingly effective tool for enabling compositional generalization.

\section{Prototype-based sequence models for data recombination}
\label{sec:prototypes}

We begin with a brief review of existing prototype-based sequence models. Our presentation mostly follows the \emph{retrieve-and-edit} approach of \citet{neural-editor}, but versions of the approach in this paper could also be built on retrieval-based models implemented with memory networks \citep{miller2016key,gu2018search} or transformers \citep{khandelwal2019generalization,guu2020realm}.
The generative process described in \cref{eq:generate} implies a marginal sequence probability:
\begin{equation}\label{eq:basedist}
    p(d) = \frac{1}{|\mathcal{D}|} \sum_{d' \in \mathcal{D}} \prewrite(d \mid d'; \theta)%
\end{equation}
Maximizing this quantity over the training set with respect to $\theta$ will encourage $\prewrite$ to act as a model of \emph{valid data transformations}: To be assigned high probability, every training example must be explained by at least one other example and a parametric rewriting operation. (The trivial solution where $p_\theta$ is the identity function, with $p_\theta(d \mid d' = d) = 1$, can be ruled out manually in the design of $p_\theta$.)
When $\mathcal{D}$ is large, the sum in \cref{eq:basedist} is too large to enumerate exhaustively when computing the marginal likelihood. Instead, we can optimize a lower bound by restricting the sum to a \textbf{neighborhood} $\mathcal{N}(d) \subset \mathcal{D}$ of training examples around each $d$:
\begin{equation}
    p(d) \geq \frac{1}{|\mathcal{D}|} \sum_{d' \in \mathcal{N}(d)} \prewrite(d \mid d'; \theta) ~ .\\
\end{equation}
The choice of $\mathcal{N}$ is discussed in more detail in \cref{sec:neighborhoods}. Now observe that:
\begin{align}
    \log p(d) &\geq  \log \bigg( |\mathcal{N}(d)| \sum_{d' \in \mathcal{N}(d)} \frac{1}{|\mathcal{N}(d)|} \prewrite(d \mid d' ; \theta) \bigg) -\log |\mathcal{D}| \\
    &\geq  \frac{1}{|\mathcal{N}(d)|}\sum_{d' \in \mathcal{N}(d)} \log \prewrite(d \mid d'; \theta) + \log \left(\frac{|\mathcal{N}(d)|}{|\mathcal{D}|}\right) \label{eq:lowerbound}
\end{align}
where the second step uses Jensen's inequality. If all $|\mathcal{N}(d)|$ are the same size, maximizing this lower bound on log-likelihood is equivalent to simply maximizing 
\begin{equation}
\label{eq:seq2seq}
     \sum_{d' \in \mathcal{N}(d)} \log \prewrite(d \mid d'; \theta)
\end{equation}
over $\mathcal{D}$---this is the ordinary conditional likelihood for a string transducer \citep{ristad1998learning} or sequence-to-sequence model \citep{sutskever2014sequence} with examples $d, d' \in \mathcal{N}(d)$.\footnote{
Past work also includes a continuous latent variable $z$,
defining:
\begin{equation}
    \prewrite(d \mid d') = \mathbb{E}_{z \sim p(z)} [\prewrite(d \mid d', z; \theta)]
\end{equation}
As discussed in \cref{sec:datasetsexps}, the use of a continuous latent variable appears to make no difference in prediction performance for the tasks in this paper. The remainder of our presentation focuses on the simpler model in \cref{eq:seq2seq}.
}

We have motivated prototype-based models by arguing that $\prewrite$ learns a model of transformations licensed by the training data. 
However, when generalization involves complex compositions, we will show that neither a basic RNN implementation of $\prewrite$ or a single prototype is enough; we must provide the learned
rewriting model with a larger inventory of parts
and encourage reuse of those parts as faithfully as possible. This motivates the two improvements on the prototype-based modeling framework described in the remainder of this section: generalization to multiple prototypes (\cref{sec:nproto}) and a new rewriting model (\cref{sec:recomb}). 

\subsection{\texorpdfstring{$n$}{n}-prototype models}
\label{sec:nproto}

To improve \emph{compositionality} in prototype-based models, we equip them with the ability to condition on multiple examples simultaneously. We extend the basic prototype-based language model to $n$ prototypes, which we now refer to as a \textbf{recombination} model $\precomb$:
\begin{align}
 d \sim \precomb(\cdot \mid d_{1:n}'; \theta)   %
    \quad \textrm{where} \quad d_{1:n}' \defeq (d_1', d_2', \ldots, d_n') &\sim p_\Omega(\cdot)%
    \label{eq:recomb-gen}
\end{align}

A multi-protype model may be viewed as a meta-learner \citep{thrun1998learning,santoro2016meta}: it maps from a small number of examples (the prototypes) to a distribution over new datapoints consistent with those examples. By choosing the neighborhood and implementation of $\precomb$ appropriately, we can train this meta-learner to specialize in one-shot concept learning (by reusing a fragment exhibited in a single prototype) or compositional generalization (by assembling fragments of prototypes into a novel configuration).
To enable this behavior,
we define a set of \emph{compatible prototypes} $\Omega \subset \mathcal{D}^n$ (\cref{sec:neighborhoods}) and let $p_\Omega \defeq \textrm{Unif}(\Omega)$. We update \cref{eq:lowerbound} to feature a corresponding multi-prototype neighborhood $\mathcal{N}: \mathcal{D} \to \Omega$.
The only terms that have changed are the conditioning variable and the constant term, and it is again sufficient to choose $\theta$ to optimize
    $
    \sum_{d_{1:n}' \in \mathcal{N}(d)} \log \precomb(d \mid d_{1:n}')
    $
over $\mathcal{D}$,
implementing
$\precomb$ as described next.

\subsection{Recombination networks}
\label{sec:recomb}

Past work has found that latent-variable neural sequence models often ignore the latent variable and attempt to directly model sequence marginals \citep{bowman2015generating}. When an ordinary sequence-to-sequence model with attention is used to implement $\precomb$, even in the one-prototype case, generated sentences often have little overlap \ with their prototypes \citep{weston2018retrieve}.
We describe a specific model architecture for $\precomb$ that does not function as a generic noise model, and in which outputs are primarily generated via explicit reuse of fragments of multiple prototypes, by facilitating \emph{copying} from independent streams containing prototypes and previously generated input tokens.

 \begin{figure}
 \vspace{-1.25em}
     \centering
     \includegraphics[width=0.8\textwidth]{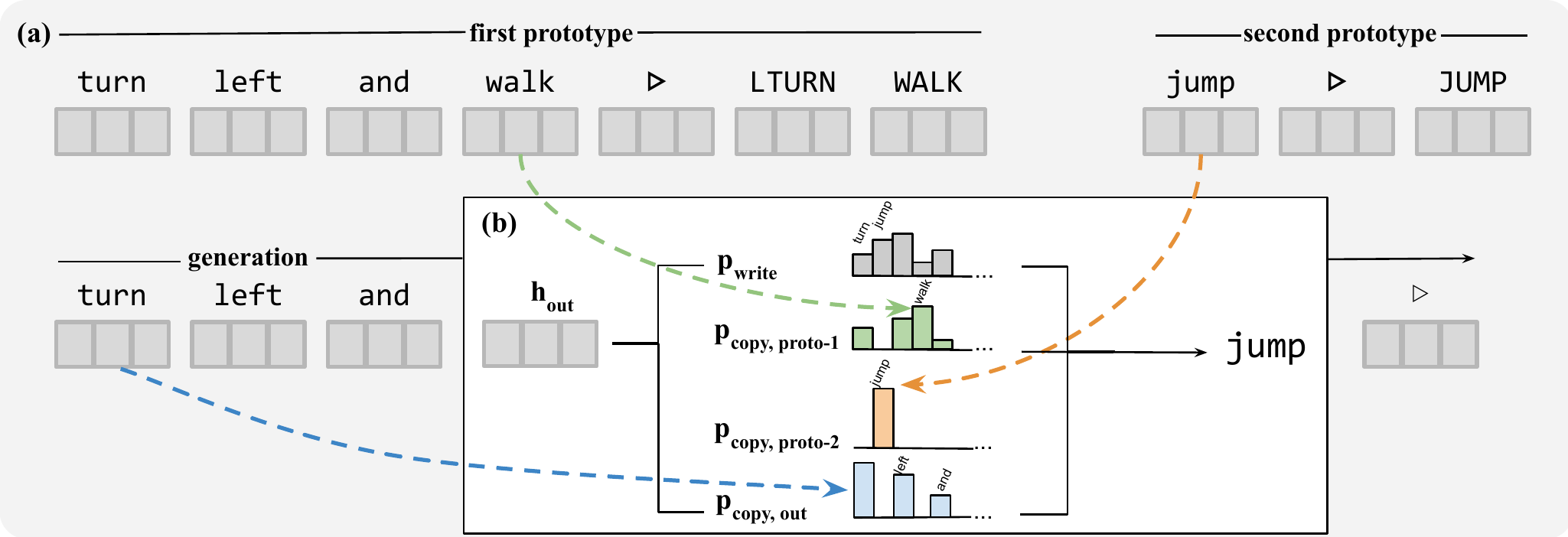}
     \caption{(a) RNN encoders produce contextual embeddings for prototype tokens. (b) In the decoder, a gated copy mechanism reuses prototypes and generated output tokens via an attention mechanism (dashed lines).}
     \label{fig:my_label}
     \vspace{-1em}
 \end{figure}

We take $\precomb(d \mid d_{1:n}'; \theta)$ to be a neural (multi-)sequence-to-sequence model \citep[c.f.][]{sutskever2014sequence}
which decomposes probability autoregressively: $\precomb(d \mid d_{1:n}'; \theta) = \prod_t p(d^t \mid d^{<t}, d_{1:n}'; \theta)$.
As shown in \cref{fig:my_label}, three LSTM encoders---two for the prototypes and one for the input prefix---compute sequences of token representations $h_\textrm{proto}$ and $h_\textrm{out}$ respectively.
Given the current decoder hidden state $h_{out}^t$,
the model first attends
to both prototype and output tokens:
\begin{alignat}{2}
    \alpha_\textrm{out}^i &\propto  \exp({h_\textrm{out}^{t^\top}} \, W_{o} \, h_\textrm{out}^i) && \quad i<t \\
    \alpha_\textrm{proto}^{kj} &\propto  \exp({h_\textrm{out}^{t^{\top}}} \, W_{p} \, h_\textrm{proto}^{kj}) && \quad k \leq n,j \leq |d_k'|
\end{alignat}
To enable copying from each sequence, we project attention weights $\alpha_\textrm{out}$ and $\alpha_\textrm{proto}^k$ onto the output vocabulary to produce a sparse vector of probabilities:
\begin{align}
p^{t}_\textrm{copy,out}(d^t = w)  
&= \textstyle \sum_{i < t} \mathds{1}[d_i = w] \cdot \alpha_\textrm{out}^i \\
p^{t}_\textrm{copy,proto-k}(d^t = w)
&= \textstyle \sum_{j \leq |d_k'|} \mathds{1}[d_{k,j}' = w] \cdot \alpha_\textrm{proto}^{kj} 
\end{align}
Unlike rule-based data recombination procedures, however, $\precomb$ is not required to copy from the prototypes, and can predict output tokens directly using \emph{values} retrieved by the attention mechanism:
\begin{align}
    h^t_\textrm{pre} &= \big[
    \textstyle
    h^t_\textrm{out}, ~~~
    \sum_i \alpha_\textrm{out}^i h^i_\textrm{out}, ~~~
    \sum_{k,j} \alpha_\textrm{proto}^{kj} h_\textrm{proto}^{kj}
    \big] \\
    p^{t}_\textrm{write} &\propto \exp(W_\textrm{write}h_\textrm{pre}^t)
\end{align}
To produce a final distribution over output tokens at time $t$, we combine predictions from each stream:
\begin{align} \label{eq:gating}
    \beta_\textrm{gate} &= \mathrm{softmax}(W_{gate}h_{out}^t) \\
    p(d^t=w \mid d^{<t}, d_{1:n}'; \theta) &= \beta_\textrm{gate} \cdot [p_\textrm{write}^t(w),p_\textrm{copy,out}^t(w),p_\textrm{copy,proto-1}^t(w),...,p_\textrm{copy,proto-n}^t(w)] %
\end{align}

This copy mechanism is similar to the one proposed by \citet{merity2016pointer} and \citet{see2017get}.
We compare 1- and 2-prototype models to an ordinary sequence model and baselines in \cref{sec:datasetsexps}.

\section{Sampling schemes}
\label{sec:neighborhoods}
\label{sec:sampling}

The models above provide generic procedures for generating well-formed combinations of training data, but do nothing to ensure that the generated samples are of a kind useful for compositional generalization. While the training objective in \cref{eq:seq2seq} encourages the learned $p(d)$ to lie close to the training data, an effective data augmentation procedure should intuitively provide \emph{novel} examples of \emph{rare} phenomena. To generate augmented training data, we combine the generative models of \cref{sec:prototypes} with a simple sampling procedure that upweights useful examples.

\subsection{Resampling augmented data} 
\label{sec:resampling}

In classification problems with imbalanced classes, a common strategy for improving accuracy on the rare class is to \emph{resample} so that the rare class is better represented in training data \citep{japkowicz2000learning}.  
When constructing an augmented dataset using the models described above,
we apply a simple rejection sampling scheme. In \cref{eq:objective}, we set:
\begin{equation}
    u(d) = \mathbb{1}[\min_t p(d^t) < \epsilon] ~ .
    \label{eq:rejection}
\end{equation}
Here $p(d^t)$ is the marginal probability that the token $d^t$ appears in any example and $\epsilon$ is a hyperparameter. The final model is then trained using \cref{eq:objective}, retaining those augmented samples for which $u(d) = 1$.
For extremely imbalanced problems, like the ones considered in \cref{sec:datasetsexps}, this weighting scheme effectively functions as a \textbf{rare tag constraint}: only examples containing rare words or tags are used to augment the original training data.

\subsection{Neighborhoods and prototype priors}

How can we ensure that the data augmentation procedure generates any samples with positive weight in \cref{eq:rejection}?
The prototype-based models described in \cref{sec:prototypes} offer an additional means of control over the generated data.\
Aside from the implementation of $\precomb$, the main factors governing
the behavior of the model are the choice of neighborhood function
$\mathcal{N}(d)$ and, for $n \geq 2$, the set of prior compatible prototypes $\Omega$.
Defining these so that rare tags also preferentially appear in prototypes helps ensure that the generated samples contribute to generalization. %
Let $d_1$ and $d_2$ be prototypes. As a notational convenience, given two sequences $d_1$, $d_2$, let $d_1 \backslash d_2$ the \emph{set} of tokens in $d_1$ but not $d_2$, and $d_1 \Delta d_2$ denote the set of tokens not common to $d_1$ and $d_2$.

\paragraph{1-prototype neighborhoods}
\citet{neural-editor} define a one-prototype $\mathcal{N}$ based on a Jaccard distance threshold \citep{jaccard1901etude}. For experiments with one-prototype models we employ a similar strategy, choosing an initial neighborhood of candidates such that
\begin{equation}
    \mathcal{N}(d) \defeq \{ d_1 \in \mathcal{D} : (\alpha \cdot |d \Delta d_1| + \beta \cdot \mathrm{lev}(d, d_1)) < \delta \}
\end{equation}
where 
$\mathrm{lev}$ is string edit distance \citep{levenshtein1966binary} and $\alpha$, $\beta$ and $\delta$ are hyperparameters (discussed in \cref{app:neighborhood}).

\paragraph{2-prototype neighborhoods}
The $n \geq 2$ prototype case requires a more complex neighborhood function---intuitively, for an input $d$, we want each $(d_1, d_2, \ldots)$ in the neighborhood to collectively contain enough information to reconstruct $d$.
Future work might treat the neighborhood function itself as
latent, allowing the model to identify groups of prototypes that make $d$ probable; here, as in existing one-prototype models, we provide heuristic 
implementations for the $n = 2$ case.

\emph{Long--short recombination}: For each $(d_1, d_2) \in \mathcal{N}(d)$, $d_1$ is chosen to be similar to $d$, and $d_2$ is chosen to be similar to the \emph{difference} between $d$ and $d_1$. (The neighborhood is
so named because one of the prototypes will generally have fewer tokens than the other one.)
\begin{align} \label{eq:longshort}
\mathcal{N}(d) \stackrel{\text { def }}{=}\{(d_1,d_2) \in \Omega :~ 
\mathrm{lev}\left(d, d_1\right) < \delta,
\mathrm{lev}\left([d\backslash d_1], d_2\right)<\delta, 
|d\backslash d_1| > 0, 
|{d\backslash d_1\backslash d_2}| = 0
\}
\end{align}
Here [$d \backslash d_1$] is the sequence obtained by removing all tokens in $d_1$ from $d$. 
Recall that we have defined $p_{\Omega}(d_{1:n}) \defeq \textrm{Unif}(\Omega)$ for a set $\Omega$ of ``compatible'' prototypes.
For experiments using long--short combination,
all prototypes are treated as compatible; that is, $\Omega = \mathcal{D} \times \mathcal{D}$.

\emph{Long--long recombination}: $\mathcal{N}(d)$ contains pairs of prototypes that are individually similar to $d$ and collectively contain all the tokens needed to reconstruct $d$: %
\begin{align} \label{eq:longlong}
\mathcal{N}(d) \stackrel{\text { def }}{=}\{(d_1,d_2) \in \Omega :~ \mathrm{lev}\left(d, d_1\right) < \delta, 
\mathrm{lev}\left(d, d_2\right)<\delta, 
|d\Delta d_1| = 1,
|{d \backslash  d_1 \backslash d_2}| = 0
\}
\end{align}

For experiments using long--long recombination, we take $\Omega = \{ (d_1, d_2) \in \mathcal{D} \times \mathcal{D} : |d_1 \Delta d_2| = 1 \}$.

\section{Datasets \& Experiments} \label{sec:datasetsexps}

We evaluate \ourmethod on two tests of compositional generalization: the \textsc{scan} instruction following task \citep{SCAN} and a few-shot morphology learning task derived from the \textsc{sigmorphon} 2018 dataset \citep{kirov-etal-2018-unimorph, sigmorphon2018}. Our experiments are designed to explore the effectiveness of learned data recombination procedures in controlled and natural settings. Both tasks involve conditional sequence prediction: while preceding sections have discussed augmentation procedures that produce data points $d = (x, y)$, learners are evaluated on their ability to predict an output $y$ from an input $x$: actions $y$ given instructions $x$, or morphological analyses $y$ given words $x$.

For each task, we compare a baseline with no data augmentation, the rule-based \textsc{geca} data augmentation procedure \citep{arxiv:geca}, and a sequence of ablated versions of \ourmethod that measure the importance of resampling and recombination. The basic \textbf{Learned Aug} model trains an RNN to generate $(x, y)$ pairs, then trains a conditional model on the original data and samples from the generative model. \textbf{Resampling} filters these samples as described in \cref{sec:sampling}. \textbf{Recomb-n} models replace the RNN with a prototype-based model as described in \cref{sec:prototypes}.
Additional experiments (\cref{tab:scan}b) compare data
augmentation to prediction of $y$ via \textbf{direct
inference} (\cref{app:direct}) in the prototype-based model and several other model variants.

\begin{table}[t!]
\vspace{-0.5em}
\centering

\caption{Results on the \textsc{scan} dataset. (a) Comparison of \ourmethod with previous work. 
Connecting lines indicate that model components are inherited from the parent (e.g.\ the row labeled \emph{recomb-2} also includes resampling).
Data augmentation with \emph{recomb-2} + resampling performs slightly worse than \textsc{geca} on the \emph{jump} and \emph{around right} splits; data augmentation with \emph{recomb-1} + resampling or an ordinary RNN does not generalize robustly to either split. All differences except between \textsc{geca} and \emph{recomb-2} + resampling in \emph{jump} are significant (paired $t$-test, $p \ll 0.001$). 
(Dashes indicate that all samples were rejected by resampling when decoding with temperature $T=1$.)
(b) Ablation experiments on the \emph{jump} split. Introducing the latent variable 
used in previous work \citep{neural-editor}  does not change performance; removing the 
copy mechanism 
results in a complete failure of generalization. While it is possible to perform conditional inference of $p(y \mid x)$ given the generative model in \cref{eq:basedist} (\emph{direct inference}), this gives significantly worse results than data augmentation (see Sec.~\ref{sec:analysis}).} 
\vspace{-0.5em}
\footnotesize
\strut
\hfill
\subfloat[][]{
\begin{tabular}{@{}>{\columncolor{white}[0pt][\tabcolsep]}ll>{\columncolor{white}[\tabcolsep][0pt]}l@{}}
\toprule
{} & \multicolumn{1}{c}{\textit{around right}} &  \multicolumn{1}{c}{\textit{jump}} \\
\midrule
 baseline    &          0.00 \stderr{0.00} &          0.00 \stderr{0.00} \\
 \textsc{geca} (published)        &      0.82 \stderr{0.11} & 0.87 \stderr{0.05} \\
 \textsc{geca} (ours)  &   \bf 0.98 \stderr{0.02} &	\bf 1.00 \stderr{0.001} \\
 learned aug.\ (basic) &       0.00 \stderr{0.00}  &  0.00 \stderr{0.00} \\
\CC  \boxRight resampling &\CC   - &\CC              - \\
\CC\boxSpace\boxDownRight \emph{recomb-1} &\CC      0.17 \stderr{0.07} &\CC             - \\
\CC\boxSpace\boxRight \emph{recomb-2} &\CC      0.75 \stderr{0.14} &\CC       0.87 \stderr{0.08} \\
\CC\emph{recomb-2} (no resampling) &\CC 0.82 \stderr{0.08} &\CC   0.88 \stderr{0.07} \\
\bottomrule
\end{tabular}
}
\hfill
\subfloat[][]{
\begin{tabular}{@{}ll@{}}
\toprule
& \multicolumn{1}{c}{\textit{jump}} \\
\cmidrule{1-2}
 \emph{recomb-2} (no resampling) & \bf 0.88 \stderr{0.07} \\
\cmidrule{1-2}
$+$ VAE & \bf {0.88} \stderr{0.07} \\
$+$ resampling & \bf {0.87} \stderr{0.08} \\
$-$ copying & 0.00 \stderr{0.00} \\
direct inference & 0.57 \stderr{0.05} \\
\bottomrule
\end{tabular}
}
\hfill
\label{tab:scan}
\vspace{-1.5em}
\end{table}

\subsection{\textsc{SCAN}}

\textsc{scan} \citep{SCAN} is a synthetic dataset featuring simple English commands paired with sequences of actions. 
Our experiments aim to show that \ourmethod performs well at one-shot concept learning and zero-shot generalization on controlled tasks where rule-based models succeed.
We experiment with two splits of the dataset, \emph{jump} and \emph{around right}. In the \emph{jump} split, which tests one-shot learning, the word \emph{jump} appears in a single  command in the training set but in more complex commands in the test set 
(e.g.\ {\it look and jump twice}). The \emph{around right} split \citep{SCANaroundright} tests zero-shot generalization by presenting learners with constructions like \emph{walk around left} and \emph{walk right} in the training set, but \emph{walk around right} only in the test set.

Despite the apparent simplicity of the task, ordinary neural sequence-to-sequence models completely fail to make correct predictions on \textsc{scan} test set (\cref{tab:scan}). As such it has been a major focus of research on compositional generalization in sequence-to-sequence models, and a number of heuristic procedures and specialized model architectures and training procedures have been developed to solve it \citep{russin2019compositional, gordon2020permutation,lake2019compositional,arxiv:geca}. Here we show that the generic prototype recombination procedure described above does so as well. We use long--short recombination for the \emph{jump} split and long--long recombination for the \emph{around right split}. We use a recombination network to generate 400 samples $d = (x,y)$ and then train an ordinary LSTM with attention \citep{bahdanau2018systematic} on the original and augmented data to predict $y$ from $x$. Training hyperparameters are provided in \cref{app:baselinemodel}.

\cref{tab:scan} shows the results of training these models on the \textsc{scan} dataset.\footnote{We provide results from \textsc{geca} for comparison. Our final RNN predictor is more accurate than the one used by \citet{arxiv:geca}, and training it on the same augmented dataset gives higher accuracies than reported in the original paper.} 
2-prototype recombination is essential for successful generalization on both splits. Additional ablations (\cref{tab:scan}b) show that the continuous latent variable used by \citet{neural-editor} does not affect performance, but that the copy mechanism described in \cref{sec:recomb} and the use of the \emph{recomb-2} model for data augmentation rather than direct inference are necessary for accurate prediction.

\subsection{\textsc{SIGMORPHON 2018}}

The \textsc{sigmorphon} 2018 dataset
consists of words paired with morphological analyses (\emph{lemmas}, or base forms, and tags for linguistic features like tense and case, as depicted in \cref{fig:teaser2}).
We use the data to construct a morphological \emph{analysis} task  \citep{akyurek2019morphological} (predicting analyses from surface forms) to test models' few-shot learning of new morphological paradigms.
In three languages of varying morphological complexity (Spanish, Swahili, and Turkish) we construct splits of the
data featuring a training set of 1000 examples and three test sets of 100 examples. 
One test set consists exclusively of words in the past tense, one in the future tense and one with other word forms (present tense verbs, nouns and adjectives). The training set contains exactly eight past-tense and eight future-tense examples; all the rest are other word forms.
Experiments evaluate \ourmethod's ability to efficiently learn noisy morphological rules, long viewed a key challenge for connectionist approaches to language learning \citep{rumelhart1986learning}.
As approaches may be sensitive to the choice of the eight examples from which the model must generalize, we construct five different splits per language
and use the Spanish past-tense data as a development set. As above, we use \textit{long--long} recombination with similarity criteria applied to $y$ only. We augment the training data with 180 samples from $\precomb$ and again train an ordinary LSTM with attention for final predictions. Details are provided in \cref{app:neighborhood}.

\begin{table}[t]
\centering
\footnotesize
\caption{$F_1$ score for morphological analysis on rare (\textsc{fut+pst}) and frequent (\textsc{other}) word forms. \ourmethod variants with 1- and 2-prototype recombination (shaded in grey) consistently match or outperform both a no-augmentation baseline and \textsc{geca}; \emph{recomb-1} + resampling is best overall. Bold numbers are not significantly different from the best result in each column under a paired t-test ($p<0.05$ after Bonferroni correction; nothing is bold if all differences are insignificant). The \textsc{novel} portion of the table shows model accuracy on examples whose exact tag set never appeared in the training data. (There were no such words in the test set for the Spanish \textsc{other}.) Differences between \textsc{geca} and the best \ourmethod variant (\emph{recomb-1} + resampling) are larger than in the full evaluation set. $^*$The Spanish past tense was used as a development set.}
\vspace{-.5em}
\resizebox{\textwidth}{!}{%
\begin{tabular}{@{}>{\columncolor{white}[0pt][\tabcolsep]}lllllll>{\columncolor{white}[\tabcolsep][0pt]}l@{}}
\toprule
& & \multicolumn{2}{c}{Spanish} & \multicolumn{2}{c}{Swahili} & \multicolumn{2}{c}{Turkish} \\
& & \multicolumn{1}{c}{\textsc{fut}+\textsc{pst}$^*$} & \multicolumn{1}{c}{\textsc{other}} 
&  \multicolumn{1}{c}{\textsc{fut}+\textsc{pst}} & \multicolumn{1}{c}{\textsc{other}} 
& \multicolumn{1}{c}{\textsc{fut}+\textsc{pst}} & \multicolumn{1}{c}{\textsc{other}} \\
\cmidrule(l){2-8}
&baseline  &       0.66 \stderr{0.01} &    0.88 \stderr{0.01} &       0.75 \stderr{0.02} &   0.90 \stderr{0.01} &         0.69 \stderr{0.04} &   0.85 \stderr{0.03} \\
&\boxRight resampling    &       0.65 \stderr{0.01} &   0.88 \stderr{0.01} &       0.77 \stderr{0.01} &   0.90 \stderr{0.02} &       0.69 \stderr{0.04} &     0.84 \stderr{0.04} \\
&\textsc{geca}  &       0.66 \stderr{0.01} &   0.88 \stderr{0.01} &       0.76 \stderr{0.02} &   0.90 \stderr{0.02} &       0.69 \stderr{0.02} &   0.87 \stderr{0.01} \\
& \boxRight resampling     &   0.72 \stderr{0.02} &  0.88 \stderr{0.01} &       0.81 \stderr{0.02} &   0.89 \stderr{0.01} &        0.75 \stderr{0.03} &   0.85 \stderr{0.02} \\
& learned aug.\ (basic)         &       0.66 \stderr{0.02} &   0.88 \stderr{0.01} &  0.77 \stderr{0.02} &   0.90 \stderr{0.01} &       0.70 \stderr{0.02} &    0.87 \stderr{0.01} \\
\rot{\rlap{\hspace{1em}\textsc{all}}}  & \CC \boxRight resampling   &\CC       0.70 \stderr{0.02} &\CC   0.86 \stderr{0.01} &\CC       \bf 0.84 \stderr{0.01} &\CC   0.90 \stderr{0.01} &\CC        0.73 \stderr{0.04} &\CC  0.85 \stderr{0.03} \\
&\CC \boxSpace\boxDownRight \emph{recomb-1}  &\CC       0.72 \stderr{0.02} &\CC   0.87 \stderr{0.01} &\CC       \bf 0.85 \stderr{0.01} &\CC   0.90 \stderr{0.02} &\CC        0.77 \stderr{0.02} &\CC   0.87 \stderr{0.02} \\
&\CC \boxSpace\boxRight \emph{recomb-2}  &\CC       0.71 \stderr{0.01} &\CC  0.87 \stderr{0.02} &\CC       0.82 \stderr{0.02} &\CC   0.90 \stderr{0.01} & \CC     0.75 \stderr{0.03} &\CC   0.86 \stderr{0.03} \\ 
&\CC   \textsc{geca} + \textit{recomb-1} + resamp.  &\CC  \bf 0.74 \stderr{0.02} &\CC    0.86 \stderr{0.01} &\CC        \bf 0.85 \stderr{0.02} &\CC    0.89 \stderr{0.01} &\CC        \bf 0.79 \stderr{0.02} &\CC    0.84 \stderr{0.01} \\
\cmidrule(l){2-8}
& baseline  &       0.63 \stderr{0.03} &       -&       0.72 \stderr{0.02} &     0.42 \stderr{0.12} &         0.68 \stderr{0.04} &     0.66 \stderr{0.15} \\
 & \textsc{geca} + resampling     &   0.67 \stderr{0.03} &      -&       0.79 \stderr{0.02} &     0.26 \stderr{0.20} &         0.73 \stderr{0.04} &     0.71 \stderr{0.10} \\
\rot{\rlap{\textsc{Novel}}} & \CC \emph{recomb-1} + resampling &  \CC    \bf 0.69 \stderr{0.02} & \CC      -& \CC     \bf 0.83 \stderr{0.02} &\CC     0.42 \stderr{0.12} &\CC      \bf  0.75 \stderr{0.03} &\CC 0.82 \stderr{0.04} \\
   &\CC  \textsc{geca} + \textit{recomb-1} + resamp. &\CC    \bf 0.69 \stderr{0.02}	  &\CC    -   &\CC \bf 0.83 	\stderr{0.02}    &\CC 	0.35 \stderr{0.11}	  &\CC \bf 0.77 \stderr{0.03} &\CC	0.71 \stderr{0.07}   \\
\bottomrule
\end{tabular}}
\label{tab:morph}
\end{table}

\cref{tab:morph} shows aggregate results across languages. We report the model's $F_1$ score for predicting morphological analyses of words in the \emph{few-shot} training condition (past and future) and the standard training condition (other word forms). Here, learned data augmentation with both one- and two-prototype models consistently matches or outperforms \textsc{geca}.
The improvement is sometimes dramatic: for few-shot prediction in Swahili, \emph{recomb-1} augmentation reduces the error rate by 40\% relative to the baseline and 21\% relative to \textsc{geca}.
An additional \textbf{baseline + resampling} experiment upweights the existing rare samples rather than synthesizing new ones; results demonstrate that recombination, and not simply reweighting, is important for generalization.
\cref{tab:morph} also includes a finer-grained analysis of \emph{novel word forms}: words in the evaluation set whose exact morphological analysis never appeared in the training set.
\ourmethod again significantly outperforms both the baseline and \textsc{geca}-based data augmentation in the few-shot \textsc{fut+past} condition and the ordinary \textsc{other} condition, underscoring the effectiveness of this approach for ``in-distribution'' compositional generalization. 
Finally, the gains provided by learned augmentation and \textsc{geca} appear to be at least partially orthogonal: combining the \textsc{geca} + resampling and \textit{recomb-1} + resampling models gives further improvements in Spanish and Turkish.

\subsection{Analysis}
\label{sec:analysis}

\paragraph{Why is \ourmethod effective?}
Samples from the best learned data augmentation models for \textsc{scan} and \textsc{sigmorphon} may be found in the \cref{app:samples} . 
We programaticaly analyzed 400 samples from \emph{recomb-2} models in \textsc{scan} and found that 40\% of novel samples are exactly correct in the \emph{around right} split and  74\% in the \emph{jump} split. A manual analysis of 50 Turkish samples indicated that only 14\% of the novel samples were exactly correct. The augmentation procedure has a high error rate! 
However, our analysis found that malformed samples either (1) feature malformed $x$s that will never appear in a test set (a phenomenon also observed by \citet{arxiv:geca} for outputs of \textsc{geca}), or (2) are mostly correct at the token level (inducing predictions with a high $F_1$ score).
Data augmentation thus contributes a mixture of irrelevant examples, \emph{label noise}---which may exert a positive regularizing effect \citep{bishop1995training}---and well-formed examples, a small number of which are sufficient to induce generalization \citep{Bastings:etal:2018}. Without resampling, \textsc{sigmorphon} models generate almost no examples of rare tags.

\paragraph{Why does \ourmethod outperform direct inference?}
A partial explanation is provided by the preceding analysis, which notes that the accuracy of the data augmentation procedure as a generative model is comparatively low. Additionally, the data augmentation procedure selects only the highest-confidence samples from the model, so the quality of predicted $y$s conditioned on random $x$s will in general be even lower. 
A conditional model trained on augmented data is able to compensate for errors in augmentation or direct inference (\cref{tab:scan-ex} in the Appendix).

\paragraph{Why is Resampling without Recombination effective?}
One surprising feature of \cref{tab:morph} is performance of the \emph{learned aug (basic) + resampling} model.
While less effective than the recombination-based models, augmentation with samples from an ordinary RNN trained on $(x, y)$ pairs improves performance for some test splits. 
One possible explanation is that resampling effectively acts as a \emph{posterior constraint} on the final model's predictive distribution, guiding it toward solutions in which rare tags are more probable than observed in the original training data. Future work might model this constraint explicitly, e.g.\ via posterior regularization \citep[as in][]{li2020posterior}.
\vspace{-.5em}

\section{Conclusions}

We have described a method for improving compositional generalization in sequence-to-sequence models via data augmentation with learned prototype recombination models. These are the first results we are aware of demonstrating that generative models of data %
are effective as data augmentation schemes in sequence-to-sequence learning problems, even when the generative models are themselves unreliable as base predictors. 
Our experiments demonstrate that it is possible to achieve compositional generalization on-par with complex symbolic models in clean, highly structured domains, and outperform them in natural ones, with basic neural modeling tools and without symbolic representations. %

\subsection*{Acknowledgments}

We thank Eric Chu for feedback on early drafts of this paper. This work was supported by a hardware donation from NVIDIA under the NVAIL grant program. The authors acknowledge the MIT SuperCloud and Lincoln Laboratory Supercomputing Center \citep{reuther2018interactive} for providing HPC resources that have contributed to the research results reported within this paper.

\bibliography{references}
\bibliographystyle{iclr2021_conference}
\newpage
\appendix
\section{Model Architecture}
\subsection{Prototype Encoder}
We use a single layer BiLSTM network to encode $h_{\textrm{proto}}^{kj}$ as follows:
\begin{align} 
    h^k_{\textrm{proto}} &= \textrm{proj}(\textrm{BiLSTM}\left(W_{e} \, d'_{k}\right))
\end{align}

\paragraph{Morphology}
 The hidden and embedding sizes are 1024. No dropout is applied. We project bi-directional embeddings to the hidden size with a linear projection. We concatenate the backward and the forward hidden states.

\paragraph{SCAN}
We choose the hidden size as 512, and embedding size as 64. We apply 0.5 dropout to the input. We project hidden vectors in the attention mechanism.

\subsection{Decoder} \label{sec:decoder}
The decoder is implemented by a single layer. In addition to the hidden state and memory cell, we also carry out a \textit{feed} vector through time:

\begin{align}
    h^t_\textrm{pre} &= \big[
    h^t_\textrm{out}, ~~~
    \sum_i \alpha_\textrm{out}^i h^i_\textrm{out}, ~~~
    \sum_{j \leq |d_1|} \alpha_\textrm{proto}^{1j} h_\textrm{proto}^{1j},~~~
    \cdots,~~~
    \sum_{j \leq |d_n|} \alpha_\textrm{proto}^{nj} h_\textrm{proto}^{nj}~~~
    \big] \\
    \textrm{feed}^t &= \textrm{Linear}_{\textrm{feed}}(h^t_\textrm{pre}) 
\end{align}

The input to the LSTM decoder at time step $t$ is the concatenation of the previous token's representation, previous feed vector, and a latent $z$ vector (in the VAE model). 

\begin{equation}
     \textrm{input}^t =  [W_d d^{t-1}, feed^{t-1}, z] \label{eq:input}
\end{equation}

\paragraph{Morphology} We use a single-layer LSTM network with a hidden size of 1024, and an embedding size of 1024. 
We initialize the decoder hidden states with the final hidden states of the BiLSTM encoder.
\emph{feed} is the same size as the hidden state.
No dropout is applied in the decoder. Output calculations are provided in the original paper in equation \cref{eq:gating}.
The query vector for the attentions is identically the hidden state: 
\begin{equation}
    \textrm{query}^t = h_{\textrm{out}}^t
\end{equation}
Further details of the attention are provided in \cref{sec:attention}.

\paragraph{SCAN}
The decoder is implemented by a single layer LSTM network with hidden size of 512, and embedding size of 64. The embedding parameters are shared with the encoder. 
Here the size of the feed vector is equal to embedding size, 64.

We have no self-attention for this decoder in the feed vector. There is an attention projection with dimension is 128. The details of the attention mechanism given in \cref{sec:attention}. 
Finally, we use transpose of the embedding matrix to project \textit{feed} to the output space. 
\begin{equation}
       \textrm{output}^t = W_e^\top \textrm{feed}^t \label{eq:output}
\end{equation}

$output^t$ contains unnormalized scores before the final softmax layer. We apply 0.7 dropout to $h_{out}^t$ during both training and test. The copy mechanism will be further described in \cref{sec:attention}.

The input to the LSTM decoder is the same as \cref{eq:input} except the decoder embedding matrix, $W_d$, shares parameters with encoder embedding matrix $W_e$. We applied 0.5 dropout to the embeddings $d_{t-1}$. 

The \textit{query} vector for the attention is calculated by:
\begin{equation}
     \textrm{query}^t =  [h_{\textrm{out}}^t, \textrm{input}^t]
\end{equation}

\subsection{Attention and Copying} \label{sec:attention}
We use the attention mechanism described in \citet{vaswani2017attention} with slight modifications.

\paragraph{Morphology}  We use a linear transformation for \textit{key} while retaining an embedding size of 1024, and leave \textit{query} and \textit{value} transformations as the identity. We do not normalize by the square root of the attention dimension.
The query vector is described in the decoder \cref{sec:decoder}. The copy mechanism for the morphology task is explained in the paper in detail.

\paragraph{SCAN}
We use the nonlinear \texttt{tanh} transformation for {\it key}, {\it query} and {\it value}. That the attention scores are calculated separately for each prototype using different parameters as well as the normalization i.e. obtaining $\alpha$'s is performed separately for each prototype. 

The copy mechanism for this task is slightly different and follows \citet{copy-mechanism-gu} We normalize prototype attention scores and output scores jointly. Let $\bar{\alpha}_i$ represent attention weights for each prototype sequence \emph{before} normalization. Then, we concatenate them to the output vector in \cref{eq:output}.
\begin{align*}
    \textrm{final}^t &= [\textrm{output}^t,\bar \alpha_1, ...,  \bar \alpha_n]
\end{align*}
We obtain a probability vector via a final softmax layer:
\begin{align*}
    \textrm{prob}^t &= \textrm{softmax}(final^t)
\end{align*}
That size of this probability vector is vocabulary size plus the total length all prototypes. We then project this into the output space by:
\begin{align*}
    p^t(w) &= \textrm{prob}^t(\texttt{indices}(w)) 
\end{align*}
where \texttt{indices} finds all corresponding scores in $prob^t$ for token $w$ where there might be more than one element for a given $w$. This is because one score can come from the $output^t$ region, and others from the prototype regions of $prob^t$. During training we applied 0.5 dropout to the indices from $output^t$. Thus, the model is encouraged to copy more.

\section{Neighborhoods and Sampling} \label{app:neighborhood}
In the \cref{eq:longshort} and \cref{eq:longlong} we expressed the generic form of neighborhood sets. Here we provide the implementation details.

\paragraph{SCAN} In the \textit{jump} split, we use long-short recombination with $\delta=0.5$. In \textit{around right} we use long-long recombination with $\delta=0.5$, and construct $\Omega$ so that the first and second prototypes to differ by a single token. We randomly pick $k < 10 \times 3$ (10 different first prototypes, and 3 different second prototypes for each of them) prototype pairs that satisfy these conditions. For the \emph{recomb-1} experiment, we use the same neighborhood setup except but consider only the $k<10$ first prototypes. 

\paragraph{Sampling} In the \textit{jump} split, we used beam search with beam size 4 in the decoder. We calculate the mean and standard deviation over the lengths of both among the first $d'_1$ and the second $d'_2$ prototypes in the train set. Then, during the sampling, we expect the first and second prototypes whose length is shorter than their respective mean plus standard deviation. This decision is based on the fact that the part of the $\Omega$ that the model is exposed to is determined by the empirical distribution, $\hat \Omega$, that arises from training neighborhoods. When sampling, we try to pick prototypes from a distribution that are close to properties of that empirical distribution.
In \emph{around right}, we use temperature sampling with $T=0.4$. 
If a model cannot sample the expected number of both novel and unique samples within a reasonable time, we increase temperature $T$.

\paragraph{Morphology}
We use long-long recombination, as explained in the paper, with slight modifications which leverage the structure of the task. We set $\Omega$ as:
$$\Omega = \{( d_1,d_2) \, | \, d'_{1\textrm{tags}} \neq d'_{2tags}, (d_{\textrm{tags}} \backslash d_{1\textrm{tags}}' \backslash d'_{2\textrm{tags}})=0\}$$
For the \emph{recomb-1} model $\mathcal{N}(d)$ utilizes tag similarity, lemma similarity and is constructed using a score function:

\begin{equation}
    \textrm{score}_1(d,d'_1) = (|d_{\textrm{tags}} \Delta d_{1\textrm{tags}}'|, \textrm{jaccard}(d_{\textrm{lemma}}, d_{1\textrm{lemma}}'))
\end{equation} 

Given $d$, we sort training examples by using $score_1$ as the comparison key and pick the four smallest neighbors (using a lexicographic sort) to form $\mathcal{N}(d)$.

For the \emph{recomb-2} model, $\mathcal{N}(d)$ uses the same score function for the first prototype as in the \emph{recomb-1} case. The second prototype is selected using:

\begin{align}
    \textrm{score}_2(d,d'_1,d'_2) = &\big(d_{\textrm{tags}} \neq d'_{2\textrm{tags}}, %
                  |d'_{1\textrm{tags}} \Delta d'_{2\textrm{tags}}|\big)
                  \label{cond}
\end{align} 

Given $x$, and a scored first prototype, we do one more sort over training examples by using $\textrm{score}_2$ as the comparison key. Then we pick first four neighbors for $\mathcal{N}(d)$.

\paragraph{Sampling}  We use a mix strategy of temperature sampling with $T=0.5$ and greedy sampling in which we use the former for $d_{\textrm{input}}$ and the latter for $d_{\textrm{output}}$. We sample 180 unique and novel examples. %

\section{Generative Model Training}\label{app:generativemodel} \paragraph{Morphology} All of the hyper parameters mentioned here are optimized by a grid search on the Spanish validation set. We train our models for 25 epochs\footnote{When training 2-proto and 1-proto models, we increment epoch counter when the entire neighborhood for every $d$ is processed. For 0-proto, one epoch is defined canonically i.e. the entire train set.}. We use Adam optimizer with learning rate 0.0001. The generative model is trained on morphological reinflection order ($d_{\textrm{lemma}} d_{\textrm{tags}} \triangleright d_{\textrm{inflection}}$) from left to right, then the samples from the model are reordered for morphological analysis task ($d_{\textrm{inflection}} \triangleright d_{\textrm{lemma}} d_{\textrm{tags}}$). 

\paragraph{SCAN} We use different number of epochs for \emph{jump} and \emph{around right} splits where all models are trained for 8 epochs in the former and 3 epochs in the latter. We use Adam optimizer with learning rate 0.002, and gradient norm clip with 1.0.

\section{Seq2Seq Baseline Model}\label{app:baselinemodel}
After generating novel samples, we either concatenate them to the training data (in morphology), or 
sample training batches from a mixture of the original training data and the augmented data. Our conditional model is the same as the generative model used in morphology experiments, described in detail in the paper body, replacing $d_{\textrm{proto}}$ with $x$, and $d$ with $y$. %

Every conditional model's size is the same as the corresponding generative model which was used for augmentation. This is to ensure that the conditional model and the generative model have the same capacity. We train conditional models for 150 epochs for SCAN and we used augmentation ratios of $p_{\textrm{aug}}=0.01$ and $p_{\textrm{aug}}=0.2$ in \textit{jump} and \textit{around right}, respectively. %
For morphology, we train the conditional models for 100 epochs, and we use all generated examples for augmentation.
\section{Direct Inference}\label{app:direct}
To adapt the prototype-based model for conditional prediction, we condition the neighborhood function on the input $x$ rather than the full datum $d$, as in \citet{hashimoto2018retrieve}. Candidate $y$s are then sampled from the generative model given the observed $x$ while marginalizing over retrieved prototypes. Finally, we re-rank these candidates via \cref{eq:seq2seq} and output the highest-scoring candidate.

\section{VAE Model}
\paragraph{Prior $p(z)$:} We use the same prior as \citet{neural-editor} given in \cref{eq:prior}.
In this prior, $z$ is defined by a norm and direction vector. The norm is sampled from the uniform distribution between zero and a maximum possible norm $\mu_{\textrm{max}}=10.0$, and the direction is sampled uniformly from the unit hypersphere. This sampling procedure corresponds to a von Mises--Fisher distribution with concentration parameter zero.

\begin{equation}\label{eq:prior}
z = z_{\textrm{norm}} \cdot z_{\textrm{dir}} \quad \textit{where} \quad z_{\textrm{norm}} \sim U(0, \mu_{\textrm{max}}) , \quad z_{\textrm{dir}} \sim  \textrm{vmF}(\vec{u},0)
\end{equation}

For SCAN, the size of $z$ is 32, and for morphology the size of $z$ is 2.

\paragraph{Proposal Network $q(z|d, d_{1:n})$:} Similarly to the prior, the posterior network decomposes $z$ into its norm and direction vectors. The norm vector is sampled from a uniform distribution at $(|\mu|, \textrm{min}(|\mu|+\epsilon, \mu_{\textrm{max}}))$, and the direction is sampled from the von Mises--Fisher distribution $\textrm{vmF}(\mu,\kappa)$ where $\kappa=25, \epsilon=1.0$. 

\begin{align} 
    h_{\textrm{final}} &= (\stackrel{\leftharpoondown}{h}_{\textrm{proto}})_{\textrm{start}} + (\stackrel{\rightharpoonup}{h}_{\textrm{proto}})_{end} \nonumber\\
    \mu &= \tanh \left(W_{z} \, \left[h_{d\textrm{final}},h_{q\textrm{final}}\right]\right) \nonumber\\
    z_{\textrm{norm}} &\sim U(|\mu|, min(|\mu|+\epsilon, \mu_{\textrm{max}}))\label{line:znorm} \\
    z_{\textrm{dir}} &\sim vmF(\mu,\kappa)\label{line:zdir} \\
    z &= z_{\textrm{norm}} \cdot z_{\textrm{dir}} \label{line:z}
\end{align}

\section{Additional Results}
\subsection{Morphology Results}
In the paper, \cref{tab:morph} shows morphology results for (non-VAE) models with 8 \emph{hints} (past- and future-tense examples in the training set). Here, we provide additional results for different hint set sizes and model variants.

\subsubsection{hints=4}

\begin{table}[H]
\centering
\caption{Exact Match Accuracy}
\resizebox{0.8\textwidth}{!}{
\begin{tabular}{lllllll}
\toprule
& \multicolumn{2}{c}{Spanish} & \multicolumn{2}{c}{Swahili} & \multicolumn{2}{c}{Turkish} \\
& \multicolumn{1}{c}{\textsc{fut}+\textsc{pst}$^*$} & \multicolumn{1}{c}{\textsc{other}} 
& \multicolumn{1}{c}{\textsc{fut}+\textsc{pst}} & \multicolumn{1}{c}{\textsc{other}} 
& \multicolumn{1}{c}{\textsc{fut}+\textsc{pst}} & \multicolumn{1}{c}{\textsc{other}} \\
\midrule
     baseline &        0.078 \stderr{0.029} &      0.63 \stderr{0.09} &        0.107 \stderr{0.034} &    0.532 \stderr{0.029} &        0.067 \stderr{0.020} &      0.57 \stderr{0.04} \\
         geca &        0.072 \stderr{0.019} &      0.63 \stderr{0.05} &        0.039 \stderr{0.011} &    0.496 \stderr{0.027} &        0.052 \stderr{0.014} &      0.54 \stderr{0.08} \\
   geca + resampling &          0.16 \stderr{0.04} &      0.65 \stderr{0.05} &          0.27 \stderr{0.08} &      0.52 \stderr{0.04} &          0.12 \stderr{0.04} &    0.554 \stderr{0.029} \\
       learned aug &        0.063 \stderr{0.012} &      0.65 \stderr{0.04} &        0.066 \stderr{0.034} &      0.52 \stderr{0.04} &        0.074 \stderr{0.021} &      0.57 \stderr{0.04} \\
 learned aug + resampling &        0.098 \stderr{0.021} &      0.65 \stderr{0.05} &          0.29 \stderr{0.06} &    0.480 \stderr{0.035} &        0.092 \stderr{0.029} &      0.54 \stderr{0.06} \\
       recomb-1 &        0.063 \stderr{0.017} &    0.674 \stderr{0.021} &        0.061 \stderr{0.017} &    0.520 \stderr{0.028} &        0.055 \stderr{0.021} &    0.554 \stderr{0.030} \\
 recomb-1 + resampling &          0.13 \stderr{0.04} &      0.64 \stderr{0.04} &          0.29 \stderr{0.04} &      0.48 \stderr{0.04} &          0.15 \stderr{0.04} &      0.52 \stderr{0.06} \\
       recomb-2 &        0.061 \stderr{0.010} &    0.656 \stderr{0.030} &          0.08 \stderr{0.06} &    0.524 \stderr{0.026} &        0.073 \stderr{0.019} &      0.58 \stderr{0.05} \\
 recomb-2 + resampling &        0.108 \stderr{0.021} &      0.64 \stderr{0.05} &          0.18 \stderr{0.04} &    0.542 \stderr{0.035} &        0.067 \stderr{0.026} &      0.55 \stderr{0.06} \\
\bottomrule
\end{tabular}
}
\end{table}

\begin{table}[H]
\centering
\caption{F1 Accuracy}
\resizebox{0.8\textwidth}{!}{
\begin{tabular}{lllllll}
\toprule
& \multicolumn{2}{c}{Spanish} & \multicolumn{2}{c}{Swahili} & \multicolumn{2}{c}{Turkish} \\
& \multicolumn{1}{c}{\textsc{fut}+\textsc{pst}$^*$} & \multicolumn{1}{c}{\textsc{other}} 
& \multicolumn{1}{c}{\textsc{fut}+\textsc{pst}} & \multicolumn{1}{c}{\textsc{other}} 
& \multicolumn{1}{c}{\textsc{fut}+\textsc{pst}} & \multicolumn{1}{c}{\textsc{other}} \\
\midrule
     baseline &       0.609 \stderr{0.025} &   0.873 \stderr{0.034} &       0.746 \stderr{0.013} &   0.897 \stderr{0.005} &       0.561 \stderr{0.032} &   0.867 \stderr{0.015} \\
         geca &       0.606 \stderr{0.019} &   0.871 \stderr{0.017} &       0.722 \stderr{0.018} &   0.884 \stderr{0.007} &       0.565 \stderr{0.034} &   0.856 \stderr{0.028} \\
   geca + resampling &       0.675 \stderr{0.017} &   0.870 \stderr{0.022} &       0.802 \stderr{0.023} &   0.892 \stderr{0.010} &         0.65 \stderr{0.05} &   0.850 \stderr{0.019} \\
       learned aug &       0.597 \stderr{0.021} &   0.871 \stderr{0.024} &       0.737 \stderr{0.010} &   0.897 \stderr{0.011} &         0.58 \stderr{0.04} &   0.853 \stderr{0.026} \\
 learned aug + resampling &       0.646 \stderr{0.007} &   0.872 \stderr{0.028} &       0.826 \stderr{0.013} &   0.887 \stderr{0.010} &       0.637 \stderr{0.032} &   0.835 \stderr{0.034} \\
       recomb-1 &       0.596 \stderr{0.020} &   0.884 \stderr{0.010} &       0.727 \stderr{0.012} &   0.893 \stderr{0.010} &       0.557 \stderr{0.032} &   0.868 \stderr{0.010} \\
 recomb-1 + resampling &       0.663 \stderr{0.029} &   0.874 \stderr{0.014} &       0.812 \stderr{0.017} &   0.886 \stderr{0.011} &         0.67 \stderr{0.04} &     0.84 \stderr{0.04} \\
       recomb-2 &       0.598 \stderr{0.019} &   0.874 \stderr{0.007} &       0.730 \stderr{0.023} &   0.894 \stderr{0.008} &       0.581 \stderr{0.035} &   0.865 \stderr{0.024} \\
 recomb-2 + resampling &       0.658 \stderr{0.012} &   0.872 \stderr{0.015} &       0.778 \stderr{0.011} &   0.897 \stderr{0.007} &       0.609 \stderr{0.032} &   0.850 \stderr{0.029} \\
\bottomrule
\end{tabular}
}
\end{table}

\subsubsection{hints=8}
Main 8-prototype $F_1$ results are provided in the body of the paper. Here we provide exact match results and an extra set of comparisons to the VAE model.%
\begin{table}[H]
\centering
\caption{Exact Match Accuracy}
\resizebox{0.8\textwidth}{!}{
\begin{tabular}{lllllll}
\toprule
& \multicolumn{2}{c}{Spanish} & \multicolumn{2}{c}{Swahili} & \multicolumn{2}{c}{Turkish} \\
& \multicolumn{1}{c}{\textsc{fut}+\textsc{pst}$^*$} & \multicolumn{1}{c}{\textsc{other}} 
& \multicolumn{1}{c}{\textsc{fut}+\textsc{pst}} & \multicolumn{1}{c}{\textsc{other}} 
& \multicolumn{1}{c}{\textsc{fut}+\textsc{pst}} & \multicolumn{1}{c}{\textsc{other}} \\
\midrule
     baseline &        0.151 \stderr{0.017} &      0.65 \stderr{0.04} &          0.15 \stderr{0.04} &    0.554 \stderr{0.034} &          0.23 \stderr{0.06} &      0.55 \stderr{0.04} \\
         geca &        0.136 \stderr{0.030} &    0.638 \stderr{0.026} &          0.15 \stderr{0.05} &      0.55 \stderr{0.06} &          0.21 \stderr{0.05} &    0.550 \stderr{0.032} \\
   geca + resampling &        0.249 \stderr{0.034} &      0.64 \stderr{0.04} &          0.25 \stderr{0.05} &    0.532 \stderr{0.033} &          0.27 \stderr{0.07} &    0.524 \stderr{0.026} \\
       learned aug &        0.163 \stderr{0.030} &    0.652 \stderr{0.033} &          0.18 \stderr{0.05} &    0.560 \stderr{0.026} &          0.23 \stderr{0.04} &    0.548 \stderr{0.019} \\
 learned aug + resampling &        0.181 \stderr{0.026} &    0.590 \stderr{0.032} &          0.34 \stderr{0.06} &    0.552 \stderr{0.029} &          0.24 \stderr{0.04} &      0.53 \stderr{0.05} \\
       recomb-1 &        0.155 \stderr{0.018} &    0.628 \stderr{0.020} &        0.161 \stderr{0.017} &    0.560 \stderr{0.025} &          0.22 \stderr{0.04} &    0.538 \stderr{0.025} \\
 recomb-1 + resampling &        0.218 \stderr{0.032} &    0.616 \stderr{0.034} &          0.35 \stderr{0.04} &      0.53 \stderr{0.04} &          0.30 \stderr{0.04} &      0.52 \stderr{0.04} \\
       recomb-2 &        0.131 \stderr{0.028} &    0.634 \stderr{0.027} &          0.19 \stderr{0.11} &      0.56 \stderr{0.04} &          0.24 \stderr{0.05} &    0.528 \stderr{0.032} \\
 recomb-2 + resampling &        0.203 \stderr{0.035} &      0.63 \stderr{0.05} &          0.27 \stderr{0.07} &    0.552 \stderr{0.031} &          0.25 \stderr{0.05} &      0.54 \stderr{0.06} \\
\bottomrule
\end{tabular}
}
\end{table}

\begin{table}[H]
\centering
\caption{$F_1$ Accuracy (VAE model)}
\resizebox{0.8\textwidth}{!}{
\begin{tabular}{lllllll}
\toprule
& \multicolumn{2}{c}{Spanish} & \multicolumn{2}{c}{Swahili} & \multicolumn{2}{c}{Turkish} \\
& \multicolumn{1}{c}{\textsc{fut}+\textsc{pst}$^*$} & \multicolumn{1}{c}{\textsc{other}} 
& \multicolumn{1}{c}{\textsc{fut}+\textsc{pst}} & \multicolumn{1}{c}{\textsc{other}} 
& \multicolumn{1}{c}{\textsc{fut}+\textsc{pst}} & \multicolumn{1}{c}{\textsc{other}} \\
\midrule
   learned aug + resampling +vae &       0.689 \stderr{0.018} &   0.859 \stderr{0.010} &       0.845 \stderr{0.014} &   0.896 \stderr{0.011} &       0.730 \stderr{0.032} &   0.850 \stderr{0.015} \\
   recomb-1 + resampling +vae &       0.717 \stderr{0.014} &   0.870 \stderr{0.007} &       0.843 \stderr{0.014} &   0.898 \stderr{0.010} &       0.736 \stderr{0.030}  &   0.859 \stderr{0.031} \\
   recomb-2 + resampling +vae &       0.710 \stderr{0.008} &   0.865 \stderr{0.012} &       0.824 \stderr{0.015} &   0.896 \stderr{0.011} &       0.751 \stderr{0.027} &   0.848 \stderr{0.027} \\
\bottomrule
\end{tabular}
}
\end{table}

\subsubsection{hints=16}
\begin{table}[H]
\centering
\caption{Exact Match Accuracy}
\resizebox{0.8\textwidth}{!}{
\begin{tabular}{lllllll}
\toprule
& \multicolumn{2}{c}{Spanish} & \multicolumn{2}{c}{Swahili} & \multicolumn{2}{c}{Turkish} \\
& \multicolumn{1}{c}{\textsc{fut}+\textsc{pst}$^*$} & \multicolumn{1}{c}{\textsc{other}} 
& \multicolumn{1}{c}{\textsc{fut}+\textsc{pst}} & \multicolumn{1}{c}{\textsc{other}} 
& \multicolumn{1}{c}{\textsc{fut}+\textsc{pst}} & \multicolumn{1}{c}{\textsc{other}} \\
\midrule
     baseline &          0.27 \stderr{0.05} &      0.65 \stderr{0.04} &          0.28 \stderr{0.06} &    0.544 \stderr{0.029} &          0.40 \stderr{0.04} &    0.614 \stderr{0.032} \\
         geca &          0.26 \stderr{0.06} &      0.65 \stderr{0.06} &          0.26 \stderr{0.05} &    0.530 \stderr{0.028} &          0.37 \stderr{0.05} &    0.570 \stderr{0.035} \\
   geca + resampling &          0.34 \stderr{0.05} &      0.63 \stderr{0.04} &          0.32 \stderr{0.07} &    0.506 \stderr{0.034} &          0.42 \stderr{0.05} &    0.590 \stderr{0.035} \\
       learned aug &          0.25 \stderr{0.04} &      0.65 \stderr{0.04} &          0.32 \stderr{0.06} &    0.538 \stderr{0.028} &          0.39 \stderr{0.04} &      0.58 \stderr{0.05} \\
 learned aug + resampling &        0.230 \stderr{0.035} &      0.61 \stderr{0.04} &          0.42 \stderr{0.06} &      0.54 \stderr{0.04} &          0.42 \stderr{0.05} &    0.578 \stderr{0.027} \\
       recomb-1 &          0.27 \stderr{0.05} &      0.63 \stderr{0.06} &          0.32 \stderr{0.05} &      0.55 \stderr{0.04} &          0.35 \stderr{0.06} &      0.60 \stderr{0.05} \\
 recomb-1 + resampling &          0.28 \stderr{0.04} &      0.61 \stderr{0.07} &        0.418 \stderr{0.035} &    0.548 \stderr{0.023} &          0.35 \stderr{0.06} &      0.56 \stderr{0.04} \\
       recomb-2 &          0.22 \stderr{0.06} &      0.62 \stderr{0.07} &          0.28 \stderr{0.04} &      0.56 \stderr{0.04} &          0.40 \stderr{0.06} &    0.596 \stderr{0.024} \\
 recomb-2 + resampling &        0.262 \stderr{0.025} &      0.61 \stderr{0.07} &        0.405 \stderr{0.028} &      0.53 \stderr{0.04} &          0.43 \stderr{0.06} &      0.61 \stderr{0.04} \\
\bottomrule
\end{tabular}
}
\end{table}

\begin{table}[H]
\centering
\caption{F1 Accuracy}
\resizebox{0.8\textwidth}{!}{
\begin{tabular}{lllllll}
\toprule
& \multicolumn{2}{c}{Spanish} & \multicolumn{2}{c}{Swahili} & \multicolumn{2}{c}{Turkish} \\
& \multicolumn{1}{c}{\textsc{fut}+\textsc{pst}$^*$} & \multicolumn{1}{c}{\textsc{other}} 
& \multicolumn{1}{c}{\textsc{fut}+\textsc{pst}} & \multicolumn{1}{c}{\textsc{other}} 
& \multicolumn{1}{c}{\textsc{fut}+\textsc{pst}} & \multicolumn{1}{c}{\textsc{other}} \\
\midrule
     baseline &       0.733 \stderr{0.014} &   0.881 \stderr{0.012} &       0.811 \stderr{0.018} &   0.893 \stderr{0.011} &       0.750 \stderr{0.026} &   0.875 \stderr{0.021} \\
         geca &       0.736 \stderr{0.019} &   0.884 \stderr{0.018} &       0.800 \stderr{0.024} &   0.889 \stderr{0.012} &         0.74 \stderr{0.04} &   0.863 \stderr{0.019} \\
   geca + resampling &       0.782 \stderr{0.024} &   0.867 \stderr{0.012} &       0.830 \stderr{0.021} &   0.885 \stderr{0.013} &       0.794 \stderr{0.032} &   0.865 \stderr{0.018} \\
       learned aug &       0.738 \stderr{0.020} &   0.877 \stderr{0.008} &       0.816 \stderr{0.024} &   0.893 \stderr{0.012} &       0.752 \stderr{0.024} &   0.868 \stderr{0.020} \\
 learned aug + resampling &       0.745 \stderr{0.019} &   0.870 \stderr{0.012} &       0.866 \stderr{0.016} &   0.894 \stderr{0.013} &       0.787 \stderr{0.031} &   0.863 \stderr{0.021} \\
       recomb-1 &       0.738 \stderr{0.021} &   0.877 \stderr{0.019} &       0.820 \stderr{0.018} &   0.896 \stderr{0.014} &       0.735 \stderr{0.033} &   0.874 \stderr{0.026} \\
 recomb-1 + resampling &       0.770 \stderr{0.020} &   0.867 \stderr{0.023} &       0.872 \stderr{0.005} &   0.892 \stderr{0.010} &       0.778 \stderr{0.024} &   0.861 \stderr{0.022} \\
       recomb-2 &       0.716 \stderr{0.019} &   0.876 \stderr{0.022} &       0.815 \stderr{0.017} &   0.897 \stderr{0.016} &       0.752 \stderr{0.034} &   0.873 \stderr{0.017} \\
 recomb-2 + resampling &       0.765 \stderr{0.023} &   0.868 \stderr{0.021} &       0.856 \stderr{0.015} &   0.888 \stderr{0.016} &       0.808 \stderr{0.018} &   0.868 \stderr{0.027} \\
\bottomrule
\end{tabular}
}
\end{table}

\subsection{Significance Tests}
Tables \ref{tab:pvalsturkish}, \ref{tab:pvalsspanish} and \ref{tab:pvalsswahili} sho the $p$-values for pairwise differences between the baseline and prototype-based models 

\begin{table}[H]
\centering
\caption{Turkish language $p$-values for paired $t$-test in PST+FUT tenses for the average $F_1$ (micro) scores over several runs without Bonferronni correction.}
\label{tab:pvalsturkish}
\resizebox{\textwidth}{!}{
\begin{tabular}{llllllllll}
\toprule
{} &     baseline &         geca &       learned aug &       recomb-1 &       recomb-2 & geca + resampling & learned aug + resampling & recomb-1 + resampling & recomb-2 + resampling \\
\midrule
baseline     &              &              &              &              &              &            &              &              &              \\
geca         &     0.259314 &              &              &              &              &            &              &              &              \\
learned aug       &     0.352506 &     0.802058 &              &              &              &            &              &              &              \\
recomb-1       &     0.707534 &     0.129244 &     0.187597 &              &              &            &              &              &              \\
recomb-2       &     0.233578 &    0.0230554 &    0.0331794 &     0.363375 &              &            &              &              &              \\
geca + resampling   &   1.0125e-16 &   3.7044e-12 &  4.07678e-15 &  6.04788e-17 &  1.71167e-19 &            &              &              &              \\
learned aug + resampling &  8.00807e-10 &  1.37553e-07 &  6.51548e-08 &  6.35314e-11 &  3.76501e-13 &  0.0167999 &              &              &              \\
recomb-1 + resampling &  3.85877e-26 &  1.60117e-20 &  2.76421e-22 &  6.41228e-26 &  2.07776e-26 &  0.0109365 &    1.948e-06 &              &              \\
recomb-2 + resampling &  2.56689e-15 &   3.4083e-13 &  2.08177e-14 &  2.92113e-18 &  1.79928e-19 &   0.981886 &    0.0190462 &    0.0101878 &              \\
\bottomrule
\end{tabular}
}
\end{table}

\begin{table}[H]
\centering
\caption{Spanish language $p$-values for paired $t$-test in PST+FUT tenses for the average $F_1$ (micro) scores over several runs without Bonferronni correction.}
\label{tab:pvalsspanish}
\resizebox{\textwidth}{!}{
\begin{tabular}{llllllllll}
\toprule
{} &     baseline &         geca &       learned aug &       recomb-1 &       recomb-2 &   geca + resampling & learned aug + resampling & recomb-1 + resampling & recomb-2 + resampling \\
\midrule
baseline     &              &              &              &              &              &              &              &              &              \\
geca         &     0.394748 &              &              &              &              &              &              &              &              \\
learned aug       &     0.761129 &     0.635337 &              &              &              &              &              &              &              \\
recomb-1       &     0.428606 &     0.974851 &     0.620601 &              &              &              &              &              &              \\
recomb-2       &     0.199768 &     0.601998 &     0.317494 &     0.625078 &              &              &              &              &              \\
geca + resampling   &  2.27478e-25 &  6.11513e-29 &  3.30904e-24 &  1.38242e-24 &  2.19894e-27 &              &              &              &              \\
learned aug + resampling &  1.09224e-10 &  9.34816e-13 &  1.40474e-11 &  4.70624e-13 &  4.78418e-14 &  0.000137083 &              &              &              \\
recomb-1 + resampling &  4.00039e-27 &  8.88347e-30 &  4.06546e-25 &   1.2159e-28 &   7.2465e-29 &     0.495734 &  1.35727e-05 &              &              \\
recomb-2 + resampling &  1.17709e-17 &  1.29429e-21 &  4.07864e-18 &  1.12332e-19 &  1.66638e-21 &     0.313143 &   0.00925477 &     0.103819 &              \\
\bottomrule
\end{tabular}
}
\end{table}

\begin{table}[H]
\centering
\caption{Swahili language $p$-values for paired $t$-test in PST+FUT tenses for the average $F_1$ (micro) scores over several runs without Bonferronni correction.}
\label{tab:pvalsswahili}
\resizebox{\textwidth}{!}{
\begin{tabular}{llllllllll}
\toprule
{} &     baseline &         geca &       learned aug &       recomb-1 &       recomb-2 &   geca + resampling & learned aug + resampling & recomb-1 + resampling & recomb-2 + resampling \\
\midrule
baseline     &              &              &              &              &              &              &              &              &              \\
geca         &     0.606002 &              &              &              &              &              &              &              &              \\
learned aug       &  0.000857131 &   0.00384601 &              &              &              &              &              &              &              \\
recomb-1       &  6.27581e-05 &   0.00101351 &     0.769589 &              &              &              &              &              &              \\
recomb-2       &  1.75947e-05 &  0.000207507 &     0.263242 &     0.402064 &              &              &              &              &              \\
geca + resampling   &  2.58696e-21 &  8.85433e-19 &  4.57259e-11 &  1.33673e-11 &  1.87968e-08 &              &              &              &              \\
learned aug + resampling &  7.09377e-53 &  1.46895e-47 &   2.3846e-38 &  6.32242e-38 &  2.09274e-30 &  3.26321e-10 &              &              &              \\
recomb-1 + resampling &  1.66361e-58 &  1.28557e-54 &  9.46035e-44 &  2.05703e-45 &  9.46848e-37 &  1.60241e-17 &    0.0749463 &              &              \\
recomb-2 + resampling &  2.16531e-31 &  3.52334e-25 &  7.80047e-20 &  1.08762e-19 &  1.26218e-15 &    0.0756646 &  1.52594e-06 &  2.51776e-11 &              \\
\bottomrule
\end{tabular}
}
\end{table}

\subsection{Generated Samples} \label{app:samples}

All samples are randomly selected unless otherwise indicated.

\subsubsection{\textsc{SCAN}}
In \cref{tab:scan-ex}, we present three test samples from the \textsc{SCAN} task along with the predictions by direct inference and the conditional model trained on the augmented data with \emph{recomb-2}. Note that the augmentation procedure was able to create novel samples whose input ($x$) happens to be in the test set (Examples 1 and 3) while $y$ may or may not be correct (Example 1). 

\begin{table}[H]
\centering
\footnotesize
\resizebox{\textwidth}{!}{
\begin{tabular}{llll}
\toprule
& Example 1 (\emph{jump}) & Example 2 (\emph{jump}) & Example 3 (\emph{around right}) \\
\midrule
Input ($x = \hat{x}$) & walk twice after jump twice & run right after jump twice & jump left and jump around right \\
True label (\textit{y}) & \texttt{JUMP JUMP WALK WALK} & \texttt{JUMP JUMP RTURN RUN} & \texttt{TURN LEFT JUMP TURN RIGHT JUMP TURN RIGHT JUMP TURN RIGHT JUMP TURN RIGHT JUMP} \\
$\hat{y}$ in augmented dataset & \textcolor{red}{\texttt{JUMP JUMP JUMP WALK}} & (not generated) & \textcolor{blue}{\texttt{TURN LEFT JUMP TURN RIGHT JUMP TURN RIGHT JUMP TURN RIGHT JUMP TURN RIGHT JUMP}}\\
Predicted \emph{$\hat{y}$}& & & \\
\boxSpace\boxDownRight direct inference & \textcolor{red}{\texttt{JUMP JUMP JUMP WALK}} & \textcolor{red}{\texttt{LOOK LOOK RTURN JUMP}} & \textcolor{red}{\texttt{TURN LEFT JUMP TURN LEFT JUMP TURN LEFT JUMP TURN LEFT JUMP TURN LEFT JUMP}}\\
\boxSpace\boxRight \emph{recomb-2}  & \textcolor{blue}{\texttt{JUMP JUMP WALK WALK}}& \textcolor{blue}{\texttt{JUMP JUMP RTURN RUN}} & \textcolor{blue}{\texttt{TURN LEFT JUMP TURN RIGHT JUMP TURN RIGHT JUMP TURN RIGHT JUMP TURN RIGHT JUMP}}\\

\bottomrule
\end{tabular}
}
\caption{Comparison of generative and unconditional model predictions with and without data augmentation. The conditional model trained on augmented data (final row) is able to compensate for errors in data augmentation (Example 1) and generalize to examples never generated by the data augmentation procedure (Example 2), and often times if a sample is correctly created by the augmentation procedure, the conditional model also gets it right (Example 3).}
\label{tab:scan-ex}
\end{table}

Below are a set of samples from the learned aug (basic) model for \textsc{SCAN} dataset's \emph{jump} and \emph{around right} splits, in order:

\begingroup
    \fontsize{8pt}{8pt}\selectfont
    \texttt{IN: run opposite and walk opposite right twice OUT: RUN TURN RIGHT TURN RIGHT RUN TURN RIGHT TURN RIGHT WALK \newline
IN: look around right thrice after run around thrice thrice OUT: TURN RIGHT RUN TURN RIGHT RUN TURN RIGHT RUN TURN RIGHT RUN TURN RIGHT RUN TURN RIGHT RUN TURN RIGHT RUN TURN RIGHT RUN TURN RIGHT RUN TURN RIGHT RUN TURN RIGHT RUN TURN RIGHT RUN TURN RIGHT LOOK TURN RIGHT LOOK TURN RIGHT LOOK TURN RIGHT LOOK TURN RIGHT LOOK \newline
IN: look opposite right twice and walk around twice OUT: TURN RIGHT TURN RIGHT LOOK TURN RIGHT TURN RIGHT LOOK TURN RIGHT WALK TURN LEFT WALK TURN LEFT WALK TURN LEFT WALK \newline
IN: run opposite and thrice OUT: RUN TURN LEFT RUN RUN \newline \newline
IN: walk opposite right thrice turn turn right thrice OUT: TURN RIGHT TURN RIGHT TURN RIGHT TURN RIGHT TURN RIGHT TURN RIGHT TURN RIGHT TURN RIGHT TURN RIGHT TURN RIGHT WALK \newline
IN: jump opposite right twice jump look around left OUT: TURN RIGHT TURN RIGHT JUMP TURN LEFT TURN LEFT JUMP TURN LEFT TURN LEFT LOOK TURN LEFT TURN LEFT LOOK \newline
IN: walk around left thrice after jump left left OUT: TURN LEFT WALK TURN LEFT WALK TURN LEFT WALK TURN LEFT WALK TURN LEFT WALK TURN LEFT WALK TURN LEFT WALK TURN LEFT WALK TURN LEFT WALK TURN LEFT WALK TURN LEFT WALK TURN LEFT WALK TURN LEFT WALK TURN LEFT WALK TURN LEFT WALK TURN LEFT WALK TURN LEFT WALK \newline
IN: run opposite right twice walk run left thrice OUT: TURN RIGHT TURN RIGHT RUN TURN RIGHT TURN RIGHT WALK TURN LEFT TURN LEFT RUN TURN LEFT TURN LEFT RUN
    }
\endgroup 

Below are a set of samples from the \emph{recomb-1} model for \textsc{SCAN} dataset's \emph{around right} split. Note that there were no samples with rare tags generated by the model for the \emph{jump} split:

\begingroup
    \fontsize{8pt}{8pt}\selectfont
    \texttt{IN: run around right after walk around left OUT: TURN LEFT WALK TURN LEFT WALK TURN LEFT RUN TURN LEFT RUN TURN LEFT WALK TURN LEFT WALK TURN LEFT WALK TURN LEFT WALK \newline
IN: look around right after jump around left OUT: TURN LEFT LOOK TURN LEFT JUMP TURN LEFT LOOK TURN LEFT LOOK TURN LEFT JUMP TURN LEFT JUMP TURN LEFT JUMP TURN LEFT JUMP \newline
IN: look around right and jump around left OUT: TURN RIGHT LOOK TURN RIGHT LOOK TURN RIGHT LOOK TURN LEFT JUMP TURN LEFT JUMP TURN LEFT JUMP TURN LEFT JUMP TURN LEFT JUMP \newline
IN: walk around right and turn right twice OUT: TURN RIGHT WALK TURN RIGHT WALK TURN RIGHT WALK TURN RIGHT WALK TURN RIGHT TURN RIGHT
    }
\endgroup 

Below are 4 samples from the \emph{recomb-2} model for each of \textsc{SCAN} dataset's \emph{jump} and \emph{around right} splits, respectively:

\begingroup
    \fontsize{8pt}{8pt}\selectfont
    \texttt{IN: jump opposite left thrice after jump opposite left thrice OUT: TURN LEFT TURN LEFT JUMP TURN LEFT TURN LEFT JUMP TURN LEFT TURN LEFT JUMP TURN LEFT TURN LEFT WALK TURN LEFT TURN LEFT WALK TURN LEFT TURN LEFT WALK \newline
IN: jump left thrice and jump left thrice OUT: TURN LEFT LOOK TURN LEFT LOOK TURN LEFT LOOK TURN LEFT JUMP TURN LEFT JUMP TURN LEFT JUMP
IN: jump opposite right and turn around left OUT: TURN RIGHT TURN RIGHT JUMP \newline TURN LEFT TURN LEFT TURN LEFT TURN LEFT \newline
IN: turn around left and jump around left OUT: TURN LEFT TURN LEFT TURN LEFT TURN LEFT TURN LEFT JUMP TURN LEFT JUMP TURN LEFT JUMP TURN LEFT JUMP \newline \newline
IN: look right twice after run around right OUT: TURN RIGHT RUN TURN RIGHT RUN TURN RIGHT RUN TURN RIGHT RUN TURN RIGHT LOOK TURN RIGHT LOOK \newline
IN: turn right twice after look around right OUT: TURN RIGHT LOOK TURN RIGHT LOOK TURN RIGHT LOOK TURN RIGHT LOOK TURN RIGHT TURN RIGHT \newline
IN: look twice and run around right OUT: LOOK LOOK TURN RIGHT RUN TURN RIGHT RUN TURN RIGHT RUN TURN RIGHT RUN \newline
IN: walk opposite right twice and jump around right OUT: TURN RIGHT TURN RIGHT WALK TURN RIGHT TURN RIGHT WALK TURN RIGHT JUMP TURN RIGHT JUMP TURN RIGHT JUMP TURN RIGHT JUMP
        }
\endgroup

\subsubsection{\textsc{Morphology}}

Below are a set of samples from the learned aug (basic) model in \textsc{sigmorphon} format.%

\begingroup
    \fontsize{8pt}{8pt}\selectfont
    \texttt{şahmiçe	şahmiçende	N;LOC;SG;PSS2S \newline
karadan havaya füze	karadan havaya füzel	N;DAT;PL;PSS3P \newline
ernek	erneklerine	N;DAT;PL;PSS3P \newline
kiler	kilerime	N;DAT;SG;PSS1S \newline
mahlep	mahlebimizi	N;ACC;SG;PSS1P \newline
süzmek	süzerler	V;IND;3;PL;PRS;POS;DECL \newline
âlap	âlaps	N;LGSPEC1;3S;SG;PRS \newline
jöle	jöleleri	N;ACC;PL \newline \newline
envejecerse	envejeciéndose	V.CVB;PRS \newline
colaxar	colaxa	V;IND;PRS;3;SG \newline
pergedrer	no pergedremos	V;NEG;IMP;1;PL \newline
mantear	no mantees	V;NEG;IMP;2;SG \newline
flaguear	no flagueen	V;NEG;IMP;3;PL \newline
malacostar	malacostaría	V;COND;3;SG \newline
desinstar	desinse	V;POS;IMP;3;SG \newline
concretizar	no concretices	V;NEG;IMP;2;SG
    }
\endgroup 

Below are a set of samples from the learned aug (basic) + resampling model.

\begingroup
    \fontsize{8pt}{8pt}\selectfont
    \texttt{şaşırmak	şaşırmıyor musun?	V;IND;2;PL;PST;PROG;POS;INTR \newline 
ayılmak	ayılmaya	V;IND;1;SG;PST;DECL \newline
pleşmek	pleşmiyor muyuz?	V;IND;1;PL;PST;PROG;NEG;INTR \newline
imciyetmek	imciyetmezdeğiz	V;IND;1;PL;FUT;NEG;DECL \newline
kuvaşmak	kuvaşmayacağız	V;IND;1;PL;FUT;NEG;DECL \newline
yermek	yermeyeceğiz	V;IND;1;PL;FUT;NEG;DECL \newline
yarıtmak	yarıtmayacağız	V;IND;1;PL;FUT;NEG;DECL \newline
kelimek	kelimeyeceğiz	V;IND;1;PL;FUT;NEG;DECL \newline \newline
trasescar	trasescáis	V;IND;PST;2;PL;IPFV \newline
tronar	tronar	V;IND;FUT;1;SG \newline
terzcalminar	terzcalminan	V;IND;PST;3;PL;IPFV \newline
esubronizar	esubronizamos	V;IND;PST;1;PL;IPFV \newline
urdir	urdiremos	V;IND;FUT;1;PL \newline
conder	conderemos	V;IND;FUT;1;PL \newline
florear	florearían	V;IND;PST;3;PL;LGSPEC1;SG \newline
sabrordar	sabrordamos	V;IND;PST;1;PL;IPFV 
    }
\endgroup

Below are a set of samples from the \emph{recomb-1} + resampling model (the best performing model in \cref{tab:morph}). Here we additionally annotate samples with error categories.

\begingroup
    \fontsize{8pt}{8pt}\selectfont
    \texttt{kovulmak    kovulmaz mısınız   V;IND;2;PL;FUT;NEG;INTR  (Inflection and tags don't match.) \newline
    düşünmek    düşündüler  V;IND;3;PL;PST;POS;DECL (Correct and novel.) \newline
    sütmek sütmez miyiz?   V;IND;2;SG;FUT;NEG;INTR (Inflection and tags don't match.) \newline
    bakmak  bakmayacak mıyım?   V;IND;1;PL;FUT;NEG;INTR (Inflection and tags don't match)\newline
    döndürmek   döndürecek misiniz? V;IND;2;PL;FUT;POS;INTR (Correct and novel.) \newline
    türkçeleştirtmek    türkçeleştirtiyor m V;IND;2;PL;PST;PROG;NEG;INTR (Wrong inflection, novel tag.) \newline
    çalmak  çalmayız    V;IND;2;PL;FUT;POS;DECL (Inflection and tags don't match.) \newline
    üsürmek    üsürmezsin  V;IND;2;SG;PST;NEG;DECL (Inflection and tags don't match.) \newline \newline
    duplicar    duplicaráis V;IND;FUT;2;PL (Correct and novel) \newline
    efundar efundan V;SBJV;FUT;3;PL (Inflection and tags don't match) \newline
    deshumanizar    deshumanicas    V;SBJV;PST;2;SG (Inplausible inflection.)\newline
    emular  emulares    V;SBJV;FUT;2;SG (Correct and also in train set.)\newline
    languidecer languidecíamos  V;IND;PST;1;SG;IPFV (Inflection and tags don't match) \newline
    nominar nominamos   V;SBJV;FUT;1;PL (Novel tags, incorrect inflection.) \newline
    finciar finciare    V;SBJV;FUT;1;SG (Correct and novel.) \newline
    abastar abasto  V;IND;PST;1;SG (Inflection and tags don't match)
    }
\endgroup 

\newpage
\subsection{Attention Heatmap}
Here we provide a visualization copy and attention mechanism in \emph{recomb-2} model for SCAN experiments.
\begin{figure}[H]
    \centering
    \includegraphics[width=\textwidth]{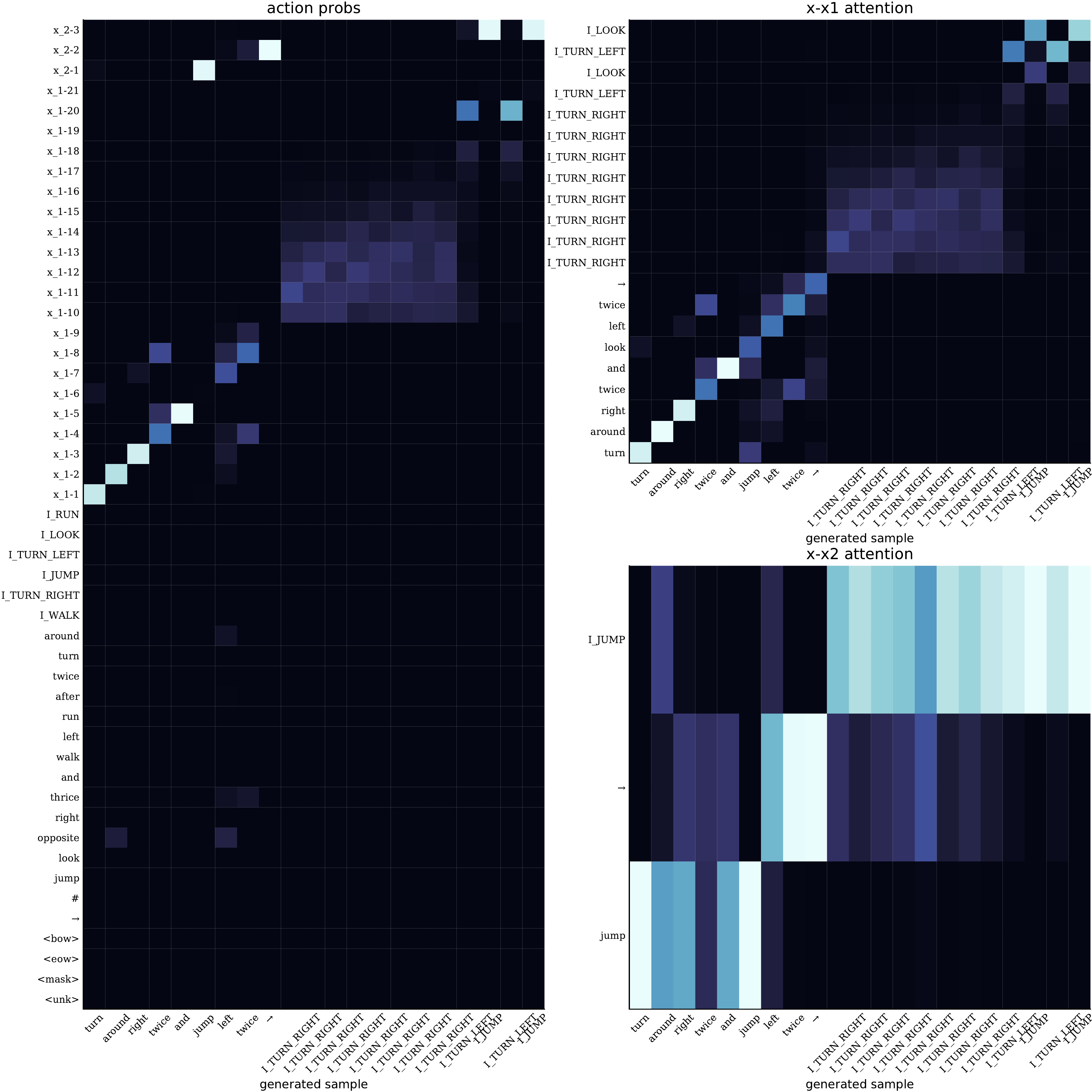}
    \caption{Generation of a sample. We plot normalized output scores on the left, and attention weights to the different prototypes on the right. The prototypes are on the y axes. The model is recomb-2 model trained on SCAN \emph{jump} split.}
    \label{tab:attentionheatmap}
\end{figure}

\section{Compute}
We use a single 32GB NVIDIA V100 Volta GPU for each experiment. For every experiment, the whole pipeline which consists of training of the generative model, sampling and training of the conditional model takes less than an hour.

\end{document}